%% file: mainNotebook.tex
\documentclass{egpubl}
\usepackage{STAG2025P}
 
\WsPaper           

\usepackage[T1]{fontenc}
\usepackage{dfadobe}  

\usepackage[toc,page]{appendix}
\usepackage[export]{adjustbox}
\usepackage{booktabs}
\usepackage{colortbl}
\usepackage{pifont}
\usepackage{subcaption}
\usepackage{multirow}
\usepackage[acronym]{glossaries}
\usepackage{overpic}
\usepackage{multirow, makecell}
\usepackage{pifont}
\usepackage{threeparttable}
\usepackage{wrapfig}
\usepackage{amssymb}

\usepackage[utf8]{inputenc}
\usepackage[T1]{fontenc}   
\newcommand{\mycomment}[1]{}
\usepackage{makecell}
\usepackage{algorithm}
\usepackage{algpseudocode}
\usepackage{xcolor}
\usepackage{tikz}
\usetikzlibrary{3d,calc,arrows.meta}
\usepackage{pgfplots}
\pgfplotsset{compat=1.18}
\usepackage{enumerate}

\newtheorem{example}{Example}

\usepackage{cite}  
\BibtexOrBiblatex
\electronicVersion
\PrintedOrElectronic

\newcommand{\y}[1]{\gls{#1}}
\newcommand{\yp}[1]{\glspl{#1}}
\newcommand{\rom}[1]{\uppercase\expandafter{\romannumeral #1\relax}}
\loadglsentries{acronyms}

\usepackage{egweblnk} 

\title{Generalizing Shape-from-Template to Topological Changes}

\author{%
  \parbox{\textwidth}{\centering
    Kevin Manogue$^{1,\ast}$\orcid{0000-0001-5015-8568}, Tomasz M Schang$^{2,\ast}$\orcid{0009-0008-7575-5201}, Dilara Kuş$^{3}$\orcid{0009-0000-4509-3130}, 
    Jonas Müller$^{4}$\orcid{0009-0003-0959-8889}, Stefan Zachow$^{5}$\orcid{0000-0001-7964-3049}, 
    Agniva Sengupta$^{3,5}$\orcid{0000-0003-0948-5641}
  }\\
  \parbox{\textwidth}{\centering
    $^1$Università degli Studi di Padova, Italy, \quad
    $^2$University of California, Berkeley, USA, \quad
    $^3$Freie Universität Berlin, Germany,\\
    $^4$Technische Universität Berlin, Germany, \quad
    $^5$Zuse Institute Berlin, Germany\\
    $\ast$Equal contribution
  }
}
\date{} 

\usepackage[noabbrev]{cleveref}
\begin{document}


\maketitle
\begin{abstract}
   Reconstructing the surfaces of deformable objects from correspondences between a 3D template and a 2D image is well studied under \y{sft} methods; however, existing approaches break down when topological changes accompany the deformation. We propose a principled extension of \y{sft} that enables reconstruction in the presence of such changes. Our approach is initialized with a classical \y{sft} solution and iteratively adapts the template by partitioning its spatial domain so as to minimize an energy functional that jointly encodes physical plausibility and reprojection consistency. We demonstrate that the method robustly captures a wide range of practically relevant topological events including tears and cuts on bounded 2D surfaces, thereby establishing the first general framework for topological-change-aware \y{sft}. Experiments on both synthetic and real data confirm that our approach consistently outperforms baseline methods.



\printccsdesc   
\end{abstract}  

\section{Introduction}
\input{sections/intro}

\section{Background}\label{sec_bg}
\input{sections/background}

\section{Method}
\input{sections/method}

\section{Results}\label{sec_results}
\input{sections/results}

\section{Conclusion}

\input{sections/conclusion}

\section*{Acknowledgements}
We thank the Graduate-Level Research in Industrial Projects for Students (G-RIPS) program at the Research Campus MODAL located at the Zuse Institute Berlin (ZIB) and the Institute for Pure and Applied Mathematics (IPAM) at the University of California, Los Angeles (UCLA). We also thank the Berlin University Alliance (BUA) for supporting this work through the Student Research Opportunities Program (StuROPx). Finally, we thank David Barac for providing the \textit{owl data}.

\bibliographystyle{eg-alpha-doi} 
\bibliography{references}

\end{document}

%% file: acronyms.tex
\newacronym{sft}{S\textit{f}T}{Shape-from-Template}
\newacronym{nrsfm}{NRS\textit{f}M}{Non-Rigid Structure-from-Motion}
\newacronym{sd}{SD}{Standard-Deviation}
\newacronym{slam}{SL\textit{a}M}{Simultaneous Localisation and Mapping}
\newacronym{icp}{ICP}{Iterative Closest Point}
\newacronym{svd}{SVD}{Singular Value Decomposition}
\newacronym{lls}{LLS}{Linear Least-Squares}
\newacronym{lm}{LM}{Levenberg-Marquardt}
\newacronym{bnb}{BnB}{Branch-and-Bound}
\newacronym{sfm}{S\textit{f}M}{Structure-from-Motion}
\newacronym{socp}{SOCP}{Second-Order Cone Programming}
\newacronym{sdp}{SDP}{Semi-Definite Programming}
\newacronym{sos}{S\textit{o}S}{Sum-of-Squares}
\newacronym{arap}{ARAP}{As-Rigid-As-Possible}
\newacronym{gt}{G\textit{t}}{Groundtruth}
\newacronym{sl}{SL}{Sight-Line}
\newacronym{po}{PO}{Polynomial Optimisation}
\newacronym{lmi}{LMI}{Linear Matrix Inequality}
\newacronym{pop}{P\textit{o}P}{Polynomial Optimisation Problem}
\newacronym{pmi}{PMI}{Polynomial Matrix Inequality}
\newacronym{llr}{L\textit{h}LR}{Lasserre's hierarchy of LMI Relaxations}
\newacronym{pnp}{P\textit{n}P}{Perspective-$n$-Point}
\newacronym{p3p}{P3P}{Perspective-3-Point}
\newacronym{ar}{AR}{Augmented Reality}
\newacronym{au}{\textit{au}}{Arbitrary Units}
\newacronym{ssm}{SSM}{Statistical Shape Model}
\newacronym{psd}{PSD}{Positive Semi-Definite}
\newacronym{pca}{PCA}{Principal Component Analysis}
\newacronym{admm}{ADMM}{Alternating Direction Method of Multipliers}
\newacronym{rms}{RMSE}{Root Mean Squared Error}
\newacronym{dof}{D\textit{o}F}{Degree-of-Freedom}
\newacronym{gpa}{GPA}{Generalised Procrustes Analysis}
\newacronym{tps}{TPS}{Thin Plate Spline}
\newacronym{lbw}{LBW}{Linear Basis Warp}
\newacronym{cd}{CD}{Chamfer Distance}
\newacronym{ct}{CT}{Computed Tomography}
\newacronym{ipm}{IPM}{Interior Point Method}
\newacronym{mcp}{MC\textit{p}}{Motion Capture}
\newacronym{rmse}{RMSE}{Root Mean Squared Error}
\newacronym[plural=ETCs,firstplural=ETCs (etc)]{etc}{ETC}{Elementary Topological Change}

%% file: sections/intro.tex
\glsreset{sft}
The \y{sft} problem concerns the reconstruction of a deformed surface from a single perspective image, given a set of keypoint correspondences between the observed surface and an undeformed template. The template is typically modeled as a smooth 2-dimensional surface embedded in $\mathbb{R}^3$, a representation that is both natural and general: even for volumetric objects, the region visible to the camera corresponds to their outer boundary surface. Over the past decade, the \y{sft} problem has been studied extensively, and numerous formulations and solutions have been proposed \cite{bartoli2015shape, chhatkuli2016stable, fuentes2021texture, casillas2019equiareal, gallardo2020shape, parashar2019local, parashar2017isometric, orumi2019unsupervised, fuentes2022deep, kairanda2025thin}. However, existing approaches implicitly assume that the surface topology is preserved during deformation. When the object undergoes topological modifications -- such as cutting, tearing, or surgical incision\footnote{We ignore topological merging (e.g., gluing, stitching), as they can be resolved by existing methods (detailed in \cref{sec_bg}) -- seams may introduce local ambiguities but do not alter the fundamental formulation} -- these methods cease to apply. The difficulty stems from the fact that, under topological changes, the separation boundaries are unknown and often imperceptible in monocular imagery, rendering their localization and subsequent reconstruction particularly challenging. In this article, we propose a method to adapt \y{sft} to the changing topology of surfaces being cut or torn apart without being reliant on detection of separation boundaries. 

Our method begins with an initial \y{sft} solution, which is typically inaccurate in the presence of topological changes due to its inability to account for cuts or tears. We subsequently adapt the object template so as to incorporate the observed modifications, leading to accurate reconstruction despite cuts or tears. In this process, we identify structural patterns of tearing that subsume all admissible topological variations on bounded two-dimensional surfaces. The effectiveness of the approach is demonstrated on both synthetic and real datasets. These results highlight the robustness of our framework and offers a new direction for expanding applicability of \y{sft} methods. To our knowledge, this article forms the first method that solves the \y{sft} problem despite topological changes without requiring explicit additional information about separation boundaries. Such an adaptation to topological changes within \y{sft} framework opens up the method to numerous use-cases that were previously unsolvable using already existing approaches\footnote{Code and data from this article are available at: \href{https://git.zib.de/asengupta/topology-aware-sft}{git.zib.de/asengupta/topology-aware-sft}}.

\noindent \textbf{Motivation}.~Deformable objects arise naturally in a wide range of computer vision tasks, including reconstruction, tracking, \y{ar}, and \y{slam}, with applications spanning robotics~\cite{zhu2022challenges}, medical imaging~\cite{maier2013optical}, automated food processing~\cite{isachsen2021fast}, and entertainment~\cite{ray2024comprehensive, tang2024screen}. In many such settings, however, the objects of interest do not merely deform smoothly but undergo genuine topological modifications, such as tearing, fracturing, cutting, or incision. Prominent examples include: (a) surgical procedures involving incisions or resections, where accurate tracking and reconstruction are required for surgical \y{ar}~\cite{collins2020augmented, chauvet2018augmented}; (b) robotic cutting and disassembly, which are central to automated food processing, agriculture, and warehouse robotics~\cite{tang2020recognition, mitsioni2021modelling}; and (c) natural processes such as glacial calving, when tracked from aerial imagery~\cite{ryan2015uav, sledz2021applications}. As detailed later in \cref{sec_bg}, existing approaches to deformable reconstruction from monocular projective data are predicated on the assumption of topology preservation, and to date no principled framework has been proposed for the monocular reconstruction of objects undergoing such topological changes.

\noindent \textbf{Contributions}.~We offer the following contributions in this article:
\begin{itemize}
    \item We propose a framework for extending \y{sft} to handle topological changes such as cutting, tearing and incisions
    \item We formalize the notion of topological changes on surfaces and offer categorization of commonly occurring topological changes
    \item We test our proposed method extensively on synthetic and real data, thoroughly comparing against existing \y{sft} approaches to establish accuracy gains
\end{itemize}

%% file: sections/background.tex
The review of prior work is organized into two principal categories. The \textit{first} encompasses methods for reconstructing deformable objects from projective data under the assumption of preserved topology. The \textit{second} focuses on approaches designed to track topological changes directly within $\mathbb{R}^n$, given data represented in the same space. This brief survey culminates in the identification of a critical research gap -- the lack of methodologies capable of reconstructing topological changes from projective data.

\subsection{Deformable Reconstruction from Perspective Projection}
Reconstructing deformable objects from their perspective projections, given established correspondences with a template, is the \y{sft} problem. The \y{sft} problem has been addressed through a variety of modalities, including classical convex and non-convex optimization techniques incorporating physics-based, statistical, and geometric priors, as well as more recent data-driven approaches based on deep learning. In what follows, we provide a survey of both traditional and deep learning–based methodologies.

\noindent \textbf{Traditional approaches}. Early formulations of the \y{sft} problem were grounded in \y{ssm}, wherein template deformations are parameterized within a low-dimensional linear subspace and fitted to observations via classical optimization. Such models leverage linear parameterizations to capture and explain complex non-linear deformations, thereby offering a parsimonious yet expressive framework. Owing to this inherent simplicity and interpretability, \y{ssm}-based methods have remained influential well beyond their initial introduction, with successful applications continuing in recent years \cite{zia2013detailed, wang2014robust, zhu2015single, zhou20153d, sengupta2025shape}. Subsequent approaches were predicated on the assumption that the imaged object could be adequately represented as a two-dimensional surface. This led to the introduction of surface-based differential constraints such as isometry, conformality, and equiareality as deformation priors in the \y{sft} problem \cite{bartoli2015shape, chhatkuli2016stable, fuentes2021texture, casillas2019equiareal, gallardo2020shape, parashar2019local, parashar2017isometric}. These surface-based formulations enabled reconstructions under large deformations, provided the deformations adhered to the corresponding physical model. However, they fail in regimes involving non-physical distortions (e.g., severe shearing, stretching, or torsion), and are intrinsically unsuitable for articulated objects that lack a surface-based representation.

A separate family of approaches employs structural-mechanics priors to model volumetric deformations \cite{malti2015linear, sengupta2020simultaneous, sengupta2021visual, kairanda2022f}. By embedding physical constraints, these methods extend reconstruction capabilities beyond purely surface-based representations. However, the parametrization of realistic material behavior introduces strong nonlinearities into the governing equations, which in turn induce highly nonconvex energy landscapes. Consequently, optimization in these formulations is often dominated by spurious stationary points, limiting their robustness and scalability.

\noindent \textbf{Learning-based approaches}. In parallel, deep learning–based approaches to \y{sft} have seen rapid development in recent years \cite{orumi2019unsupervised, fuentes2021texture, fuentes2022deep, kairanda2025thin}. These methods demonstrate strong performance in regimes where large, domain-specific training datasets are available, effectively learning complex, nonlinear deformation priors directly from data. Nonetheless, their reliance on empirical risk minimization over a fixed training distribution introduces a fundamental limitation: performance deteriorates under distributional shifts (\emph{domain gaps}), and generalization to arbitrary input data with limited or no training supervision remains intrinsically weak. This dependence on abundant and domain-consistent training data constrains the applicability of such methods in many practical settings.

Crucially, \textit{none of the aforementioned methods are designed to explicitly accommodate topological changes} and, as our results later demonstrate (in \cref{sec_results}), they fail when confronted with data exhibiting such topological changes.

\subsection{Tracking Topological Changes}
When information about objects in $\mathbb{R}^n$ undergoing topological changes is directly available in the ambient $n$-dimensional space, one may employ established methods for tracking such changes, typically restricted to $n \in \{2,3\}$. A classical line of work exploits numerical algorithms that couple shape-gradient and topological-gradient methods within a level-set framework, enabling structural and boundary optimization with automatic hole nucleation and boundary evolution \cite{allaire2002level, allaire2005structural}. Another prominent direction introduces variational phase-field formulations, such as those developed for ductile fracture in elastic–plastic solids at finite strains, which provide a robust means of capturing crack nucleation and complex crack-topology evolution \cite{miehe2017phase}. Complementary to these are approaches based on `topological derivatives', which characterize the first-order sensitivity of suitably defined energy functionals with respect to infinitesimal shape perturbations, and thus furnish principled criteria for topology optimization \cite{sokolowski2003optimality}. More recent work has explored spline-based representations, e.g., B-spline extensions designed to directly accommodate topological changes \cite{pawar2018dthb3d_reg}. While conceptually appealing, such representations incur substantial computational costs, a difficulty that is exacerbated in projective settings, such as the typical \y{sft} framework, where composition of multiple warps must be resolved.

\subsection{Research Gap}
Although a few other methods have attempted to address topological changes from projective data \cite{colins2008automatic, tsoli2016tracking}, these invariably rely on supplementary 3D information obtained from specialized sensors, which are often unavailable in practical settings (e.g., laparoscopic or endoscopic surgery, 3D reconstruction or \y{ar} with hand-held cameras). Consequently, the problem of topological-change–aware \y{sft} remains unaddressed in the existing literature.

%% file: sections/method.tex
In this section, we present our approach to the problem of \y{sft} under topological changes, proceeding in three steps. We first review the details of the isometric \y{sft} problem, which serves as important theoretical context for our problem, and clarify why current surface-based \y{sft} methods are unable to accommodate topological changes. Following this, we introduce a novel framework for describing topological changes in a surface---beginning with elementary topological changes that model a single tear in a surface and generalizing to more complex compositions of tears and deformations. Using these notions, we formulate the \y{sft} problem with topological changes. Finally, we present our method, which optimizes a displacement vector field to correct depth values in an initializing reconstruction, and discuss implementation details.

\subsection{Notations}
We denote functions with both Greek letters, such as $\Psi$ and $\Phi,$ and unbolded Latin letters, such as $L, C,$  and $d.$ We use bold Latin letters for points, vectors, and matrices, and either Greek or Latin letters for scalars.  When defining a map $\Psi$ we use the notation $\Psi:=$ to indicate definition. We denote the Jacobian of a diffeomorphism $\Psi$ evaluated at a point $\mathbf{p}$ as $\mathbf{J}_\Psi(\mathbf{p}),$ and we denote the $n$-dimensional identity matrix as $\mathbf{Id}_n.$ We write $\mathcal{C}^n$ to refer to $n$ times continuously differentiable maps and $\mathcal{C}^\infty$  to refer to infinitely continuously differentiable maps. We consider all surfaces to be regular, and write both surfaces and their parametrization spaces using upper-case calligraphic font, such as $\mathcal{T}$ and $\mathcal{P}.$  We denote by a smooth curve $\Gamma$ in a surface $\mathcal{M}\subset \mathbb{R}^3$ the range of a $\mathcal{C}^\infty$ map $\Gamma:[0,1]\to\mathcal{M}.$ We call $\Gamma$ non-self intersecting if the map $\Gamma$ is injective on $[0,1).$ We use a tilde to indicate lifting to 3 dimensions in homogeneous coordinates, i.e. if $\mathbf{p}=(x_1,x_2)$ then $\widetilde{\mathbf{p}}=(x_1,x_2,1)$. 

 \subsection{Preliminaries}\label{sec_method_prelims}
Let $\mathcal{M}$ be a two-dimensional smooth manifold-with-boundary embedded in $\mathbb{R}^3.$ The embedding endows it with both intrinsic differential structure and extrinsic geometry relative to the ambient Euclidean space. 

\noindent\textbf{Deformation model}.~Isometric maps are those transformations that preserve distance between points along a surface. They model deformations of common materials such as paper, cloth, or some biological tissues, which can deform without being measurably stretched or compressed. Formally, isometric maps can be defined as follows.
\begin{definition}\label{def:isometry}
A diffeomorphism between two surfaces $\Psi:\mathcal{M}\to\mathcal{N}$ is called an \textit{isometry} if for all points $\mathbf{p}\in\mathcal{M}$
its Jacobian map $\mathbf{J}_\Psi(\mathbf{p}
):T_\mathbf{p}\mathcal{M}\to T_{\Psi(\mathbf{p})}\mathcal{N}$ satisfies 
\begin{equation}\label{eq:isometry}
\mathbf{J}_\Psi(\mathbf{p}){\mathbf{J}^\top_\Psi}(\mathbf{p})=\mathbf{Id}_2.
\end{equation}
\end{definition}
\Cref{eq:isometry} is measured locally, relying only on information near a given point $\mathbf{p}\in\mathcal{M}.$ As we discuss in the \cref{sft_topological}, this locality makes the above definition well-suited to the case of topological changes, unlike other, non-local definitions of isometry. For instance, geodesic preservation necessitates an exact determination of the intersection points between the geodesic and the separation boundaries. In a monocular setting, however, such localization becomes practically infeasible in the absence of additional domain-specific priors.

Isometry serves as a versatile method for modeling the unknown deformations of a known 2D template surface. While various other differential constraints, such as angle-preserving conformality \cite{bartoli2015shape} and area-preserving equiareality \cite{casillas2019equiareal}, have been studied as deformation models in the \y{sft} problem, isometry is a well-established, widely-studied model in deformable reconstruction literature \cite{huang2008non, bartoli2015shape, chhatkuli2016stable, fuentes2021texture, gallardo2020shape, parashar2017isometric, chen2022efficient}. Thus, we concentrate on the isometric case, which offers both mathematical tractability and strong empirical relevance.

\input{figures/SfT}
\noindent \textbf{Isometric \y{sft}}.~The isometric \y{sft} problem considers a connected, 2D surface template $\mathcal{T}\subset\mathbb{R}^3$ embedded in $\mathbb{R}^3$ and parametrized by a $C^1$ map $\Delta:\mathcal{P}\to\mathcal T$ from the 2D parametrization space $\mathcal{P}\subset \mathbb{R}^2$, This template is altered by an unknown isometry $\Psi:\mathcal{T}\to\mathcal{S}$ to produce the deformed surface $\mathcal{S}\subset \mathbb{R}^3$. The central problem is to determine the unknown deformation $\Psi$ given only projective data---that is, an image of the deformed surface $\mathcal{S}$. This process of estimating 3D depth data from 2D correspondences between $\mathcal{P}$ and $\mathcal{J}$ is referred to as \textit{reconstruction} in the sequel. In practice, it is often useful to work with the unknown parametrization of deformed surface $\mathcal{S},$ given by $\varphi:=\Psi\circ\Delta.$ The core elements of the \y{sft} problem are visualized in \cref{fig:SfT}. 

The restriction to a single snapshot of projective information, as opposed to wide baseline observation across images as in \y{nrsfm}~\cite{kumar2020non, sengupta2024totem} or three-dimensional data from specialized sensors~\cite{newcombe2015dynamicfusion, innmann2016volumedeform, slavcheva2017killingfusion, slavcheva2018sobolevfusion}, adds to the complexity and nuance of the \y{sft}. This difficulty is exacerbated in the presence of topological changes such as cutting or incisions. 
While modeling perspective projection is nontrivial, within the bounds of \y{sft} the camera's intrinsic parameters are assumed to be known and accounted for such that its effects correspond to perspective projection onto the $z=1$ plane in $\mathbb{R}^3$. Formally, the projective data is captured via a \textit{perspective projection function} $\Pi:\{(x,y,z)\in\mathbb{R}^3:z>0\}\to\mathbb{R}^2$ defined as $\Pi(x,y,z)=(\frac{x}{z},\frac{y}{z}).$ Thus, the \y{sft} problem is to reconstruct the unknown isometry $\Psi$ given only the projective projection $\mathcal{J}:=\Pi(\mathcal{S})\subset \mathbb{R}^2$ of the unknown deformed surface $\mathcal{S}$. In existing \y{sft} paradigm, $\mathcal{T}$ is connected and $\Psi$ is an isometry, thus the deformed surfaced $\mathcal{S}$ and the image data $\mathcal{J}$ are both connected as well, explicitly precluding topological changes. 
\input{figures/elementary-topological-changes}

\noindent \textbf{Warps}.~Surface-based \y{sft} methods assume point correspondences exist to establish a warp between the parametrization space $\mathcal{P}$ and the image data $\mathcal{J}$. Typical warps used in \y{sft} include the tensor-product B-spline \cite{Rueckert1999BSpline} and linear or radial basis warps (e.g. the \y{tps} \cite{bookstein1989principal}), the latter of which has the advantage of having non-local basis which makes it easily adaptable to topological changes (c.f., truncating local basis of B-splines are highly expensive \cite{pawar2018dthb3d_reg}). Formally, the warp appears as a $\mathcal{C}^1$ map $\eta:\mathcal{P}\to\mathcal{J}.$ We assume no occlusion occurs, meaning the projection $\Pi$ is injective and so is the warp $\eta = \Pi \circ \Psi \circ \Delta$. 


In this work, we choose to model all maps using \yp{tps} and \yp{lbw}. Both are defined as the radial basis warps with kernel $\rho: [0,\infty) \to \mathbb R$. Given control handles $\{(\mathbf{x}_i,\mathbf{y}_i)\}$, typically assigned to the keypoint correspondences, the interpolant function is defined as:
\[ f(\mathbf{x}) = \sum_i c_i\rho(|\mathbf{x}-\mathbf{x}_i|)\]
where $\{c_i\}$ are computed such that the relations $f(\mathbf{x}_i)=\mathbf{y}_i$ hold for all $i$. For \y{tps} interpolation the kernel is $\rho_{\mathrm{TPS}}(r) = r^2\ln r$, and for \y{lbw} interpolation the kernel is $\rho_{\mathrm{LBW}}(r) = r$.


\noindent \textbf{Isometric depth function}.~The warp $\eta$ allows for a practical solution to the unknowns in the \y{sft} problem. The warp $\eta = \Pi \circ \Psi \circ \Delta$ is the composition of an isometric map $\Psi$ with two non-isometric maps $\Pi$ and $\Delta$\footnote{$\Delta$ is generally non-isometric, typically obtained by quasi-conformal flattening to obtain an approximately flat parametrization \cite{sheffer2005abf++}}, while the composed parametrization $\varphi$ can be written as 
\begin{equation}\label{eq:isoSfT}
\varphi(\mathbf{p})=\widetilde \eta(\mathbf{p})\gamma(\mathbf{p}),
\end{equation}
where the map $\widetilde \eta$ is lifted to 3 dimensions in homogeneous coordinates. The scalar map $\gamma$ is referred to as the \textit{depth function} of the reconstruction $\varphi$. Isometric \y{sft} resolves an analytic form for $\gamma$ in terms of the derivatives of the known maps $\Delta$ and $\eta.$ The classical \y{sft} method in \cite{bartoli2015shape} solves the depth function as:
{\small
\begin{multline}\label{eq:depth}
\gamma(\mathbf{p})=\\
\sqrt{\lambda\left( \mathbf{J}_\Delta ^T(\mathbf{p})\mathbf{J}_\Delta(\mathbf{p})\left(\mathbf{J}_\eta^T(\mathbf{p})\mathbf{J}_\eta(\mathbf{p}) - \frac{1}{\lVert \widetilde\eta \rVert_2^2} \mathbf{J}_\eta^T(\mathbf{p})\eta (\mathbf{p})\eta^T(\mathbf{p})\mathbf{J}_\eta(\mathbf{p})\right)^{-1}\right)},
\end{multline}
}where the function $\lambda$ computes the smallest eigenvalue of its argument matrix. In this formula, it can be shown that both eigenvalues of the matrix in the argument of $\lambda$ are positive. Once $\gamma$ is found, $\Psi,$ or equivalently $\varphi,$ is recovered from \cref{eq:isoSfT} via the known maps $\eta$ and $\Delta.$

\subsection{Topological Changes}\label{sft_topological}
We define our problem setup closely following the framework established in \cref{sec_method_prelims}. Building on this setup, we introduce additional definitions that extend \y{sft} to accommodate topological changes such as tearing and discuss the integration of these concepts into isometric \y{sft}.

The warp $\eta$ is generated from point correspondences under the assumption that the image data $\mathcal{J}$ is connected. In the presence of topological changes such as cutting or tearing, these warps become ill-posed since their basis functions, defined globally, propagate influence across regions that are no longer adjacent, yielding distorted mappings inconsistent with the altered topology. As the depth function $\gamma$ is reliant on the first-order behavior of the warp (see \cref{eq:depth}), this produces incorrect depth values and broadly inaccurate reconstructions of the unknown deformation $\Psi$. Truncating the basis of warps to accommodate topological changes has been tried with B-splines \cite{giannelli2012thb, pawar2018dthb3d_reg} but with a hierarchical approach enabled by precise localization of truncation boundaries through a very expensive framework, an impractical proposition for \y{sft}. Moreover, existing surface-based isometric \y{sft} methods are theoretically inapplicable because topological changes such as tearing or fragmenting, are neither isometries nor continuous, failing to satisfy \cref{def:isometry}. Thus, in order to integrate topological changes into the framework established in \cref{sec_method_prelims}, it is necessary to introduce a broader range of transformations. 

\glsreset{etc}
\noindent \textbf{\yp{etc}}.~
Recall that the map $\Psi:\mathcal{M}\to \mathbb{R}^3$ is called a  diffeomorphism onto its range if $\Psi(\mathcal{M})$ is a regular surface and $\Psi:\mathcal{M}\to \Psi(\mathcal{M})$ is a diffeomorphism of surfaces. Similarly, with  $\mathcal{\widetilde M}\subset\mathcal{M}$ a subset of the surface $\mathcal{M},$ the restriction of $\Psi$ to $\mathcal{\widetilde M}$ is the same map defined on the restricted domain, notated as 
$\Psi|_{\mathcal{\widetilde M}}:\mathcal{\widetilde M}\to \mathbb{R}^3.$ Using these notions, we formalize the idea of a simple topological change like a tear or rip in the following definitions. 
\glsreset{etc}
\begin{definition}\label{def:tearing curve}
A smooth curve $\Gamma$ in a surface $\mathcal{M}$ is a called a \textit{regular tearing curve} if it is non-self intersecting and if it does not intersect the domain boundary $\partial\mathcal{M}$ except at the end-points $\Gamma(0)$ or $\Gamma(1).$
\end{definition}
\begin{definition}\label{def:elemtopchange}
A map $\Phi:\mathcal{M}\to\mathbb{R}^3$ is called an \y{etc} if there exists a regular tearing curve $\Gamma$ such that the restricted map $\Phi|_{\mathcal{M}\setminus\Gamma}:\mathcal{M}\setminus\Gamma\to\mathbb{R}^3$ is a diffeomorphism onto its range. If the restriction is further an isometry onto its range, $\Phi$ is called an \textit{isometry with \y{etc}}. In each case, we call $\mathcal{M}\setminus\Gamma$ the \textit{torn domain} of $\Phi$ and $\Phi(\mathcal{M}\setminus\Gamma)$ the \textit{torn range} of $\Phi.$
\end{definition}
We distinguish notation for regular diffeomorphisms and diffeomorphisms with topological changes by using $\Psi$ to refer to the former and $\Phi$ for the latter.
\begin{example}~A sheet of paper is cut into two pieces and these pieces are subjected to any composition of non-rigid and rigid transformations. The tearing curve $\Gamma$ corresponds to the path of the cut, the torn domain ${\mathcal{M}\setminus\Gamma}$ corresponds to the disconnected surfaces of two pieces of paper generated by the cut, and the transformation $\Phi|_{\mathcal{M}\setminus\Gamma}$ corresponds to the separation of the two pieces after the cut. 
\end{example}

The torn range of an \y{etc} is always a regular surface, but not necessarily connected. If $\Gamma$ is a curve that fully disconnects the surface $\mathcal{M}$, both the torn domain and torn range of $\Phi$ have two connected components. \y{etc} consists of two steps. First,  $\Gamma$ is removed from $\mathcal{M}$. Second, the resulting surface, now torn, is transformed. If the transformation of the torn surface $\mathcal{M}\setminus\Gamma$ is a diffeomorphism, then $\Phi$ is an \y{etc}. If the transformation of the torn surface $\mathcal{M}\setminus\Gamma$ is an isometry, satisfying \cref{eq:isometry} everywhere---excepting the curve $\Gamma,$ which has been deleted---then $\Phi$ is an isometry with \y{etc}. Just as isometries model physical deformations, isometries with \y{etc} model similar physical deformations after a smooth tear has been created along the path $\Gamma$. 

\noindent \textbf{Classes of \y{etc}}.~We note that the tearing curve $\Gamma$ need not begin or end on the boundary of the surface $\mathcal{M}.$ It is permitted to lie entirely within the interior of $\mathcal{M}.$ By labeling the endpoints of $\Gamma$ as $\mathbf{p}$ and $\mathbf{q}$ and the boundary of $\mathcal{M}$ as $\partial\mathcal{M},$ we identify four representative classes of \y{etc}, visualized in \cref{fig:elementary-topological-changes}, that we label as follows
\begin{enumerate}[I]
\item \textit{Exterior partial tear}: $\mathbf{p}\neq \mathbf{q}$, either $\mathbf{p}$ or $\mathbf{q}$ lies in $\partial\mathcal{M}$ but not both
\item \textit{Interior partial tear}: $\mathbf{p}\neq \mathbf{q}$, neither $\mathbf{p}$ nor $\mathbf{q}$ lies in $\partial\mathcal{M}$
\item \textit{Simple disconnection}: $\mathbf{p}\neq \mathbf{q}$ both $\mathbf{p}$ and $\mathbf{q}$ lie in $\partial\mathcal{M}$
\item \textit{Hole disconnection}: $\mathbf{p}= \mathbf{q}$, neither $\mathbf{p}$ nor $\mathbf{q}$ lies in $\partial\mathcal{M}.$
\end{enumerate}
The first two \y{etc} classes, being partial tears, preserve the global connectedness of the surface $\mathcal{M}$. While the exterior partial tear case remains simply connected, the interior partial tear case introduces nontrivial homology via the presence of an interior cut. The second two \y{etc} classes disconnect the surface $\mathcal{M}$ into two connected components. In the simple disconnection case, the resulting components are both simply connected. In the hole disconnection case, one component is simply connected and homeomorphic to a disk, while the other has non-trivial homology and is homeomorphic to an annulus. The topological properties of the torn range across the four ETC classes are collected in \cref{table:ETC}.

\begin{table}[b]
\centering
\scriptsize
\setlength{\tabcolsep}{4pt} 
\renewcommand{\arraystretch}{1.6}
\begin{tabular}{rcc}
\cline{2-3}
\cline{2-3}
 & \textbf{\# Connected components} & \textbf{Homology}\\
\hline
Exterior partial tear   & 1 & Trivial\\ 
\hline
Interior partial tear  & 1 & Non-trivial\\ 
\hline
Simple disconnection & 2 & Both trivial\\ 
\hline
Hole disconnection & 2 (Disk, Annulus) & \makecell{Disk trivial \\ Annulus non-trivial}\\ 
\bottomrule
\end{tabular}
\caption{Topological properties of the four ETC classes. The number of connected components is the zero-th Betti number of the torn range surface. The homology is measured by the first Betti number of the connected components of the torn range.}
\label{table:ETC}\end{table}

\noindent \textbf{Topological changes}.~While the four cases above require that tears be smooth and non self-intersecting, more elaborate tears can be constructed through the composition of the classes of \yp{etc}. The general definition of `topological changes' thus admits arbitrarily complicated tearing and cutting, as defined below:
\begin{definition}
A map $\Phi:\mathcal{M}\to\mathbb{R}^3$ is called a \textit{topological change} of $\mathcal{M}$ if it is the composition of finitely many \yp{etc}. Similarly, $\Phi$ is called an \textit{isometry with topological change} if it is the composition of finitely many isometries with \yp{etc}. If $\Phi$ is the topological change generated by the $k$ \yp{etc} $\Phi_1,\dots,\Phi_k$ with corresponding tearing curves 
$\Gamma_1,\dots,\Gamma_k,$ we call $\mathcal{M}\setminus\Gamma_1\cup\cdots\cup\Gamma_k$ the \textit{torn domain} of $\Phi$ and $\Phi(\mathcal{M}\setminus\Gamma_1\cup\cdots\cup\Gamma_k)$ the \textit{torn range} of $\Phi.$
\end{definition}
Where the torn range of an \y{etc} has at most two connected components, the torn range of a general topological change may have any finite number of connected components, generated by composing \yp{etc} (e.g., ripping a paper into arbitrarily many pieces). Similarly, while the tearing curve $\Gamma$ contained in an \y{etc} is required to be smooth, the tears of a general topological change need only be piecewise smooth and may feature non-smooth corners. 
\begin{example}~The torn range of a general topological change $\Phi$ consists of a finite number of connected components, each of which may feature various further partial tears, along with diffeomorphic (or isometric) deformations from the torn domain. See subfigures (e) and (f) in \cref{fig:images}, in which two printed textured papers are torn into three connected components and displaced. 
\end{example}
\input{figures/t-SfT}
\subsection{\y{sft} with Topological Changes}
In this subsection, we extend the formulation of the \y{sft} problem to incorporate topological changes, as formalized in \Cref{sft_topological}, and develop the corresponding framework for its solution.

\noindent \textbf{Problem setup}.~The setup, illustrated in \cref{fig:t-SfT}, considers a topological change represented by a map $\Phi:\mathcal{T}\to\mathbb{R}^3$, whose range is the torn surface $\mathcal{S}\subset\mathbb{R}^3$. The map $\Phi$ is assumed to be an isometry with topological change, so that the torn range $\mathcal{S}$ may in general be disconnected and consist of $k$ connected components. We label these components as $\mathcal{S}_1,\dots,\mathcal{S}_k$, and denote the corresponding components of the image data $\mathcal{J}$ as $\mathcal{J}_1,\dots,\mathcal{J}_k$. For each connected component $\mathcal{S}_i$, we write $\partial\mathcal{S}_i$ for its boundary. The objective is then to resolve the unknown \textit{isometry with topological change} $\Phi$ consistent with the observed data $\mathcal{J}_1,\dots,\mathcal{J}_k$.

We denote by $\Phi_0:\mathcal{T}\to\mathcal{S}$ the initializing reconstruction generated by the depth function \cref{eq:depth} of isometric \y{sft}. We denote the depth function of $\Phi_0$ as $\gamma_0.$  As previously discussed, the reconstruction generated by $\Phi_0$ is in general inaccurate in the presence of topological changes. In particular, as the isometric \y{sft} method \cite{bartoli2015shape} assumes global isometry of the output surface $\mathcal{S}$ and image data $\mathcal{J},$ the reconstruction generated by $\Phi_0$ typically distorts its reconstructed surface $\mathcal{S}$ across tears and separations in an attempt to maintain isometry. We now describe our method for eliminating such distortions induced by the assumption of global isometry in $\mathcal{S}$.




\noindent \textbf{Isometry deviation correction}.~We propose to correct this distorted reconstruction from $\Phi_0$ by introducing a \textit{depth displacement field} $d:\mathcal{P}\to \mathbb{R}^2$ that modifies $\Phi_0$. This displacement field produces a refined reconstruction that we label $\Phi_d,$ defined to satisfy the relation
\begin{equation}\label{eq:modified-reconstruction}
\Phi_d\big(\Delta(\mathbf{p})\big)=\varphi_d(\mathbf{p}):=\widetilde \eta(\mathbf{p})\gamma_0\big(\mathbf{p}+d(\mathbf{p})\big).
\end{equation}
At points where $d$ vanishes, $\Phi_d$ aligns with $\Phi_0.$ Where the displacement field is nonzero, it adjust the depth function of $\Phi_0$ as discussed above, giving the point $\mathbf{p}$ the depth value of a nearby point displaced from $\mathbf{p}$ in the direction $d(\mathbf{p})$.  In order to produce this displacement field, we compute $d:\mathcal P\to \mathbb{R}^2$ such that it minimizes certain cost functions defined on the torn domain. Specifically, we measure the extent by which $\Phi_0$ fails to be an isometry at a given point $\mathbf{p}\in \mathcal{M},$ by introducing the pointwise \textit{isometry error function} $L_\Phi$ of a topological change $\Phi$, defined on the torn domain as 
\begin{equation}\label{eq:isom-deviation}
    L_\Phi(\mathbf{p}) = \lVert \mathbf{J}_\Phi(\mathbf{p}){\mathbf{J}^\top_\Phi}(\mathbf{p})-\mathbf{Id}_2\rVert_{2}^2 + \lVert \mathbf{J}_{\Phi^{-1}}(\mathbf{p}){\mathbf{J}^\top_{\Phi^{-1}}}(\mathbf{p})-\mathbf{Id}_2\rVert_{2}^2.
\end{equation}
This map is symmetrized in order to capture effects from both large and small eigenvalues of the metric tensor. 
The isometry error vanishes at those points where a map $\Phi$ is isometric and positive at those points where $\Phi$ stretches, compresses, or otherwise deviates from isometry. $L_{\Phi_0}$ is near zero on the interior of the connected components of the torn domain $\Phi^{-1}(\partial \mathcal{S}_i)$, leading to more accurate reconstructions in these regions, but grows large near the boundaries of $\Phi^{-1}(\partial \mathcal{S}_i)$, where the effects of topological changes distort the depth function $\gamma_0$. This observation forms the basis of our solution method, described as follows. 

As the depth function $\gamma_0$ is primarily skewed near the separation boundaries $\partial \mathcal{S}_i$, an improved reconstruction is obtained from $\Phi_0$ by modifying the value of the depth function $\gamma_0$ near these boundaries. In particular, displacing the depth value of points near $ \partial\mathcal S_i$ to minimize deviation from isometry generates markedly better reconstructions. 

\noindent \textbf{Optimization}.~The corrective displacement field $d:\mathcal P\to \mathbb{R}^2$ minimizes the cost function
\begin{equation}\label{eq:cost}
    C(d) =  \int_{\mathcal T} \lambda L_{\Phi_d}(\mathbf{x}) + (1-\lambda) \frac{\lvert d(\Delta^{-1}(\mathbf{x}))\rvert }{ L_{\Phi_0}(\mathbf{x})+\epsilon} \mathrm d \mathbf{x}.
\end{equation}
Here, $\mathbf{x}$ denotes a point on the template $\mathcal{T}$, $\lambda \in (0,1)$ is a tunable parameter, $\epsilon$ is a small positive value, and $L_{\Phi_d}$ is the isometry error function \cref{eq:isom-deviation} adapted to $\Phi_d.$ The first term of \cref{eq:cost} serves to penalize isometry loss in the refined reconstruction $\Phi_d,$ ensuring that the displacement field points in a direction that improves the base reconstruction $\Phi_0.$ The second term of \cref{eq:cost} incentivizes larger displacement values at points where the base reconstruction $\Phi_0$ is highly non-isometric. The desired solution is $\Phi_d,$ where 
\begin{equation}
d = \operatorname*{argmin}_{d'\in\mathcal C^1(\mathcal P,\mathbb R^2)} C(d').     
\end{equation}
We solve this minimization using gradient descent. Denoting the iteration of the gradient descent by $t$, we search for $d$ by solving 
\begin{equation}
    \frac{\mathrm{d}}{\mathrm{d}t}d' = -\frac{\delta C}{\delta d'}
\end{equation}
where the right-hand side above is the functional gradient, which we estimate numerically. We initialize this functional differential equation with a random perturbation of the $\mathbf{0}$ vector field. The specific solution we find is therefore given by
\begin{equation}
    d = \lim_{t\to\infty} d'(t).
\end{equation}



\subsection{Implementation}   
We describe our implementation details below.

\noindent \textbf{Data and initialization}.~Each data set provides us with a set of $M$ points $\{\mathbf{p}_i\}\subseteq \mathcal P$ which we call \emph{source points} (in all cases, we pre-process to ensure $\mathcal P\subseteq [-1,1]^2$);
a set of $M$ points $\{\mathbf{t}_i\}\subseteq \mathcal T$ which we call \emph{template target points};
and a set of $M$ points $\{\mathbf{p}'_i\}\subseteq \mathcal{J}$ which we call \emph{image target points}.
In addition, we create: a set of $K$ \emph{displacement points} $\{\mathbf{r}_i\}\subseteq \mathcal P$, which are all the points on a $\sqrt{K}\times\sqrt{K}$ evenly spaced grid of  over the domain $[-0.95,0.95]^2$ where $K$ is given by 
\[K = \left\lceil C\sqrt{M}\right\rceil ^2\]
with $C\in[1,2]$;
a set of $K$ points $\{\mathbf{r}'_i\}\subseteq \mathcal P$ which we call \emph{displacement target points};
and a set of $N=33$ points $\{\mathbf{q}_i\}\subseteq \mathcal P$ which we call \emph{loss points}.

\noindent \textbf{Details of the warps}.~The map $\eta$ is defined by \y{lbw} or \y{tps} interpolation on the correspondences $\{(\mathbf{p}_i,\mathbf{p}'_i)\}$; the map $\Delta$ is defined by \y{tps} or \y{lbw} interpolation on the correspondences $\{(\mathbf{p}_i,\mathbf{t}_i)\}$; the map $d$ is defined by \y{tps} interpolation on the correspondences $\{(\mathbf{r}_i,\mathbf{r}'_i)\}$; the map $\varphi_d$ is defined by \y{lbw} or \y{tps} interpolation on the correspondences $\{(\mathbf{p}_i, \varphi_d(\mathbf{p}_i))\}$, where we use the modified reconstruction formula from \cref{eq:modified-reconstruction}. To estimate the loss function \cref{eq:cost}, we estimate the integral using a sum over all \emph{loss points}. To compute $\mathbf{J}_\Phi$ and $\mathbf{J}_{\Phi^{-1}}$, we treat $\Phi$ as the composition $\varphi\circ\Delta^{-1}$. Both $\varphi$ and $\Delta$ are represented by warps, so the derivatives can be interpolated and composed correctly at all \emph{loss points}.


\noindent \textbf{Gradient descent}.~For the gradient descent, we first define a \emph{step parameter} $h = 1.9/3\sqrt{K}.$
We initialize the differential equation with initial random perturbation of the zero vector field, i.e. each $\mathbf{r}'_i$ is simulated from a uniform distribution on $[-3h/10, 3h/10]^2$. 
At each iteration, we estimate the discrete derivative $D_{i,j}$ of $C(d)$ with respect to $(\mathbf{r}'_i)_j$ for each $1\leq i\leq K, 1\leq j\leq 2$. We then define $D_{\mathrm{max}}=\max_i \sqrt{D_{i,1}^2+D_{i,2}^2}$. Finally, we update all $\mathbf{r}'_i$ according to 
\[(\mathbf{r}'_i)_j \leftarrow (\mathbf{r}'_i)_j + \left(\min(1, h /2D_{\mathrm{max}})\right) D_{i,j}.\]

We take a minimum of 10 gradient descent steps, and we continue taking gradient descent steps until there are either five iterations in a row without improvement in the minimum-so-far of the loss function, or until we reach a maximum iteration count of 40. We return the displacement field and reconstruction at the iteration that minimized the loss function as the solution.

We implemented our method in Python, using the libraries \texttt{numpy}, \texttt{scipy}, and \texttt{plotly}. A simplified algorithm summarizing the many steps of our proposed approach is given in \cref{alg:topo-sft}.

\begin{mycomment}{
What to say about typical runtime?
\textcolor{blue}{Do not say anything about runtime here, we will put this in the results - but later, we will need the approx. runtime for the two real data we show below - in seconds}

In order to implement this method, we need to discretize the various quantities and maps.
\subsubsection{Setup}
\begin{enumerate}
    \item A set of $N$ points $\{\mathbf{p}_i\}\subseteq \mathcal P$ which we call \textbf{interpolation points}.
    \item A set of $M$ points $\{\mathbf{q}_i\}\subseteq \mathcal P$ which we call \textbf{source points}.
    \item A set of $M$ points $\{\mathbf{t}_i\}\subseteq \mathcal T$ which we call \textbf{template target points}.
    \item A set of $M$ points $\{\mathbf{j}_i\}\subseteq \mathcal{J}$ which we call \textbf{image target points}.
    \item A set of $K$ points $\{\mathbf{r}_i\}\subseteq \mathcal P$ which we call \textbf{displacement points}.
    \item A set of $K$ points $\Theta := \{\mathbf{\theta}_i\}\subseteq \mathcal P$ which we call \textbf{displacement target points}
\end{enumerate}  

In our specific simulations, we use the following parameters and assumptions.
\begin{enumerate}
    \item We assume $\mathcal P\subseteq [-1,1]^2$ (possible through pre-processing).
    \item We use a $\sqrt{K}\times\sqrt{K}$ evenly spaced grid of \textbf{displacement control points} over the domain $[-0.95,0.95]^2$ where $K$ is given by
    \[K = \left\lceil C\sqrt{M}\right\rceil ^2\]
    and $C_K\in [1,2]$ is chosen to ensure an efficient runtime. Higher $C$ is preferable, but takes longer. The lowest $C$ value we take is $K=3/2$.
    \item We use $N=33$, i.e. a $33\times 33$ evenly spaced grid of \textbf{interpolation points} over the domain $[-1,1]^2$.
    \item We use the \textbf{step parameter}
    \[\delta = \left(\frac{1}{3}\right)\frac{2\times 0.95}{\sqrt{K}}.\]
    \item We use the \textbf{initial condition parameter} 
    \[B=3\delta/10.\]
    \item We use the \textbf{termination parameter}
    \[S = 5.\]
    \item All warps are thin-plate spline warps.
\end{enumerate}

\subsection{Problem statement}
\subsection{Implementation}
In order to implement this method, we need to discretize the various quantities and maps.
\subsubsection{Setup}
\begin{enumerate}
    \item A set of $N$ points $\{\mathbf{p}_i\}\subseteq \mathcal P$ which we call \textbf{interpolation points}.
    \item A set of $M$ points $\{\mathbf{q}_i\}\subseteq \mathcal P$ which we call \textbf{source points}.
    \item A set of $M$ points $\{\mathbf{t}_i\}\subseteq \mathcal T$ which we call \textbf{template target points}.
    \item A set of $M$ points $\{\mathbf{j}_i\}\subseteq \mathcal{J}$ which we call \textbf{image target points}.
    \item A set of $K$ points $\{\mathbf{r}_i\}\subseteq \mathcal P$ which we call \textbf{displacement points}.
    \item A set of $K$ points $\Theta := \{\mathbf{\theta}_i\}\subseteq \mathcal P$ which we call \textbf{displacement target points}
\end{enumerate}  
The warp $\eta$ is defined on $\{\mathbf q_i\}$ by 
\[\eta(\mathbf q_i) = \mathbf j_i,\quad 1\leq i\leq M\]
and for points $\mathbf p\not\in \{\mathbf q_i\}$ it is defined by radial basis function (RBF) interpolation, for example a thin-plate spline.

Similarly, the map $\Delta$ is defined on $\{\mathbf q_i\}$ by 
\[\eta(\mathbf q_i) = \mathbf t_i,\quad 1\leq i\leq M\]
and for points $\mathbf p\not\in \{\mathbf q_i\}$ it is defined by RBF interpolation.

\subsubsection{Displacement field}
We parametrize $d=d_\Theta$ by 
\[ 
d_\Theta(\mathbf{r}_i) = \mathbf{\theta_i},\quad 1\leq i\leq K
\]
and for points $p\not\in \{\mathbf r_i\}$, it is defined by RBF interpolation.
\subsection{Reconstruction}
We parametrize $\varphi[d_\Theta]$ by defining
\[\varphi[d](q_i) = \widetilde\eta (q_i) \gamma\big(q_i + d (q_i)\big), \quad 1\leq i\leq M.\]
Again, use RBF interpolation for $p\not\in \{\mathbf q_i\}$.

\begin{enumerate}
    \item I'm not sure if the connectivity structure approach will work because of the partial tear case. Partial tear preserves connectivity of the surface. We could discuss convexity, for example, since tearing a surfaces makes it a non-convex domain. 
    \item Instead, I propose the following definition 
\end{enumerate}
\begin{definition}
Let $\Omega\subset \mathbb{R}^2$ be a closed domain with boundary $\partial\Omega.$ A map $\Psi:\Omega\to\mathbb{R}^3$ is called an \textbf{elementary topological change} if there exists a smooth, non self-intersecting curve $\Gamma\subset\mathbb{R}^2$ such that the restricted map $\Psi|_{\Omega\setminus\Gamma}:\Omega\setminus\Gamma\to\mathbb{R}^3$ is a diffeomorphism onto its image. 
If the restriction $\Psi|_{\Omega\setminus\Gamma}$ is an isometry onto its image, then $\Psi$ is called an \textbf{isometry with elementary topological change}.
\end{definition}

\begin{definition}
A map $\Psi:\Omega\to\mathbb{R}^3$ is called a \textbf{topological change} of $\Omega$ if it can be written as a composition of a finite number of elementary topological changes.
\end{definition}
Remarks:
\begin{itemize}
    \item If $\Gamma$ is taken as the empty set, the above definition corresponds to a normal isometry
    \item This definition can be generalized to $\mathbb{R}^3$ 
    \item The above definition captures our four fundamental cases. To see this, denote the endpoints of the curve $\Gamma$ as $\mathbf{p}$ and $\mathbf{q}.$ Then:
    \begin{enumerate}
        \item if $\mathbf{p}=\mathbf{q},$ we recover the (interior) cut-out  case 
        \item if $\mathbf{p}\neq \mathbf{q}$ and both points lie on interior of $\Omega,$ we recover the partial tear interior case
        \item if $\mathbf{p}\neq \mathbf{q}$ and only one lies on the boundary $\partial\Omega,$ we recover the partial tear exterior case 
        \item if $\mathbf{p}=\mathbf{q}$ and both points lie on the boundary $\partial\Omega,$ we recover the complete split case 
          
    \end{enumerate}
\end{itemize}

Specify assumptions if any...
\begin{itemize}
    \item Q: strictly speaking tearing is not a topological change. How are we defining tearing/ripping in a mathematical sense?
    \item Q: re the above, what assumptions are placing on $\Psi$? Diffeomorphic up to a set (C1 curve?) of measure zero?
\end{itemize}
Describe problem setup with a figure

\textit{Question:} should we describe the four elementary tearing operations?

    }
\end{mycomment}

%% file: figures/SfT.tex
\begin{figure}
    \centering

\begin{tikzpicture}[
            scale=1.2,
            arrow/.style={-{Latex}, thick}
        ]

\draw[dashed] (1.7,-0.9) to (1.7,2.55);

\node at (-2,0) {
    \begin{tikzpicture}
        \draw[fill=gray!50] (-1,-1) rectangle (1,1);
    \end{tikzpicture}
};


\node at (-2, 3) {
    \begin{tikzpicture}[scale=0.45]
    \begin{axis}[ticks=none,hide axis, colormap={blueblack}{color=(gray!100) color=(gray!30)}]
        \addplot3[line width=1pt, samples=31, samples y=0,domain=-1:1,variable=\t]
            ({\t},{1},{1+sin(90)});
        \addplot3[line width=2pt, samples=31, samples y=0,domain=-1:1,variable=\t]
            ({-1},{\t},{1+sin(90*\t)^2});
        \addplot3[surf,
            opacity=0.9,
            domain=-1:1,
            shader = flat,
            shader=interp,
            samples=31,
        ] {1+sin(90*y)^2};
        \addplot3[line width=1pt, samples=31, samples y=0,domain=-1:1,variable=\t]
            ({1},{\t},{1+sin(90*\t)^2});
        \addplot3[line width=1pt, samples=31, samples y=0,domain=-1:1,variable=\t]
            ({\t},{-1},{1+sin(90)});
    \end{axis}
    \end{tikzpicture}
};

\node at (1.9, 3) {
    \begin{tikzpicture}[scale=0.5]
    \begin{axis}[ticks=none,hide axis, colormap={blueblack}{color=(gray!100) color=(gray!30)}]
        \addplot3[line width=2pt, samples=31, samples y=0,domain=-1:1,variable=\t] (
            {cos(60)*( \t) - sin(60) * 1},
            {sin(60)*( \t) + cos(60) * 1},
            {3+1+0.8*sin(90 * 1)}
        );
        \addplot3[line width=2pt, samples=31, samples y=0,domain=-1:1,variable=\t] (
            {cos(60)*( \t) - sin(60) * -1},
            {sin(60)*( \t) + cos(60) * -1},
            {3+1+0.8*sin(90 * -1)}
        );
        \addplot3[surf,
            opacity=0.9,
            domain=-1:1,
            shader=interp,
            samples=31,
        ] (
        {cos(60)*(x) - sin(60) * y},
        {sin(60)*(x) + cos(60) * y},
        {3+1+0.8*sin(90 * y)}
        );
        \addplot3[line width=1pt, samples=31, samples y=0,domain=-1:1,variable=\t] (
            {cos(60)*( -1) - sin(60) * \t},
            {sin(60)*( -1) + cos(60) * \t},
            {3+1+0.8*sin(90 * \t)}
        );
        \addplot3[line width=1pt, samples=31, samples y=0,domain=-1:1,variable=\t] (
            {cos(60)*( 1) - sin(60) * \t},
            {sin(60)*( 1) + cos(60) * \t},
            {3+1+0.8*sin(90 * \t)}
        );

        \addplot3[only marks, mark=*, black] coordinates {(-0.39,0.25,3)};
    \end{axis}
    \end{tikzpicture}
};

\node at (2, 0) {
    \begin{tikzpicture}[scale=0.6]
    \begin{axis}[ticks=none,hide axis, colormap={blueblack}{color=(gray!100) color=(gray!30)}]
        \addplot3[only marks, mark=*, blue] coordinates {(0,0,0)};
        \draw[dashed,blue] (axis cs:0,0,0) -- (axis cs:-0.4,-0.8,1) ;
        \draw[dashed,blue] (axis cs:0,0,0) -- (axis cs:0.8,-0.8,1) ;
        \draw[dashed,blue] (axis cs:0,0,0) -- (axis cs:-0.4,0.6,1) ;
        \draw[dashed,blue] (axis cs:0,0,0) -- (axis cs:0.8,0.6,1) ;

        \addplot3[surf,
            opacity=0.4,
            domain=-0.4:0.8,
            y domain=-0.8:0.6,
            shader=interp,
            samples=31,
        ] (
            {x},
            {y},
            {1}
        );
        
        \addplot3[line width=2pt, samples=31, samples y=0,domain=-1:1,variable=\t] (
            {(cos(60)*( \t) - sin(60) * 1 ) / (3+1+0.8*sin(90 * 1))},
            {(sin(60)*(\t) + cos(60) * 1) / (3+1+0.8*sin(90 * 1))},
            {1}
        );
        \addplot3[line width=2pt, samples=31, samples y=0,domain=-1:1,variable=\t] (
            {(cos(60)*( \t) - sin(60) * -1) / (3+1+0.8*sin(90 * -1))},
            {(sin(60)*( \t) + cos(60) * -1) / (3+1+0.8*sin(90 * -1))},
            {1}
        );
        \addplot3[surf,
            opacity=0.9,
            domain=-1:1,
            shader=interp,
            samples=31,
        ] (
            {(cos(60)*( x) - sin(60) * y) / (3+1+0.8*sin(90 * y))},
            {(sin(60)*( x) + cos(60) * y) / (3+1+0.8*sin(90 * y))},
            {1}
        );
        \addplot3[line width=1pt, samples=31, samples y=0,domain=-1:1,variable=\t] (
            {(cos(60)*(  -1) - sin(60) * \t) / (3+1+0.8*sin(90 * \t))},
            {(sin(60)*(-1) + cos(60) * \t) / (3+1+0.8*sin(90 * \t))},
            {1}
        );
        \addplot3[line width=1pt, samples=31, samples y=0,domain=-1:1,variable=\t] (
            {(cos(60)*( 1) - sin(60) * \t) / (3+1+0.8*sin(90 * \t))},
            {(sin(60)*( 1) + cos(60) * \t) / (3+1+0.8*sin(90 * \t))},
            {1}
        );
        \addplot3[only marks, mark=*, black] coordinates {(0.065,-0.13,1)};
    \end{axis}
    \end{tikzpicture}
};

\draw[dashed] (1.7,0.33) to (1.7,1);

\node (P) at (-2.5,0.5) {$\mathcal{P}$};
\node (T) at (-2.3,3.5) {$\mathcal{T}$};
\node (S) at (1.3,3.4) {$\mathcal{S}$};
\node (J) at (3,0.4) {$\mathcal{J}$};

\draw[arrow] (-2.2, 1) to[bend left=10] node[left] {$\Delta$} (-2.2,2.2);
\draw[arrow] (-0.7, 3) to[bend left=10] node[above] {$\Psi$} (1,3);
\draw[arrow] (-1, -0.1) to[bend right=10] node[above] {$\eta$} (0.5,-0.1);
\draw[arrow] (2, 2.1)  to node[right] {$\Pi$} (2, 0.9);
\draw[arrow] (-1.3, 1) to[bend left = 10] node[above] {$\varphi$} (1,2.9);
\draw[arrow] (-1, 0.7) to[bend left = 10] node[above] {$\gamma$} (1.4,1.5);


\end{tikzpicture}

    \caption{\y{sft} \textit{without} topological changes \cite{bartoli2015shape}. The known parametrized surface $\mathcal T$ with known parametrization $\Delta$ is deformed to produce the unknown deformed surface $\mathcal{S}.$ Using the known warp $\eta$ and the known projective image data $\mathcal{J},$ \y{sft} reconstructs the unknown depth function $\gamma$ which leads to the unknown parametrization $\varphi,$ as well as the unknown deformation $\Psi$.}

    \label{fig:SfT}
\end{figure}
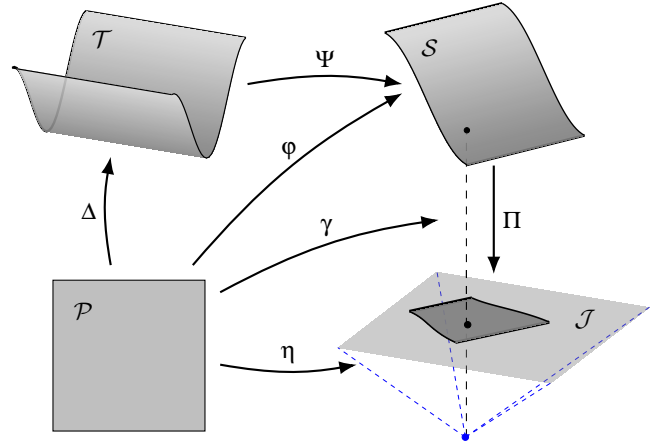

%% file: figures/elementary-topological-changes.tex
\glsunset{etc}
\begin{figure*}[t]
    \centering
    \begin{subfigure}[b]{0.24\textwidth}
        \centering
        \includegraphics[width=\linewidth]{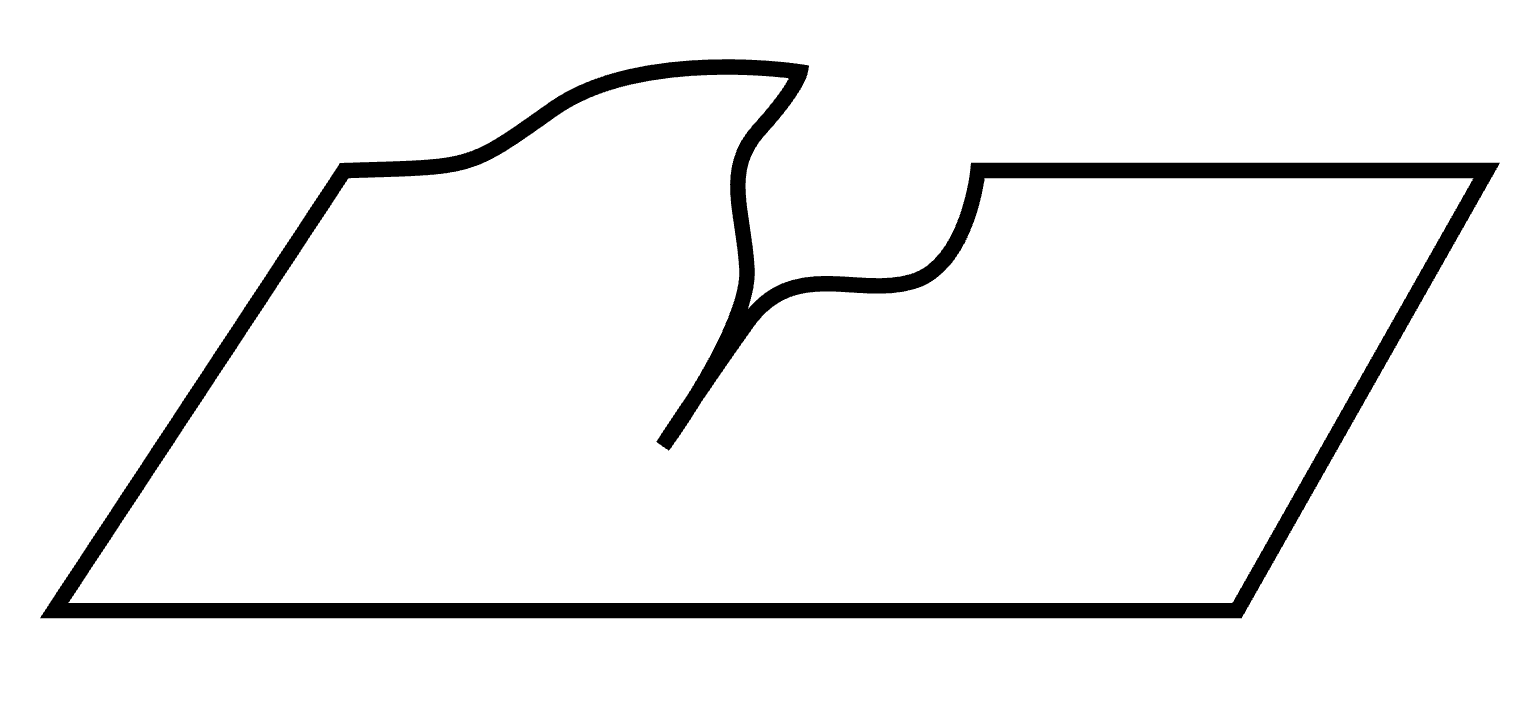}
        \caption{Exterior partial tear}
    \end{subfigure}%
    \begin{subfigure}[b]{0.24\textwidth}
        \centering
        \includegraphics[width=\linewidth]{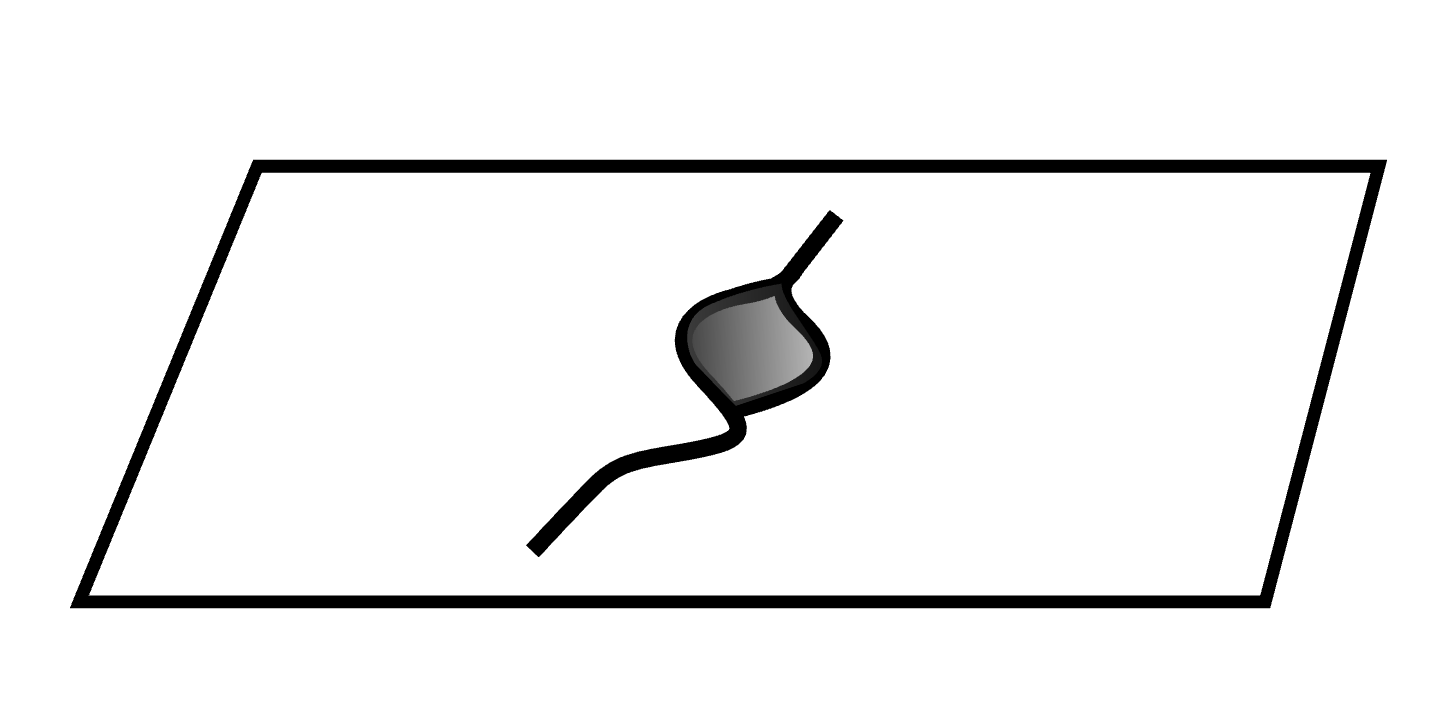}
        \caption{Interior partial tear}
    \end{subfigure}
    \begin{subfigure}[b]{0.24\textwidth}
        \centering
        \includegraphics[width=\linewidth]{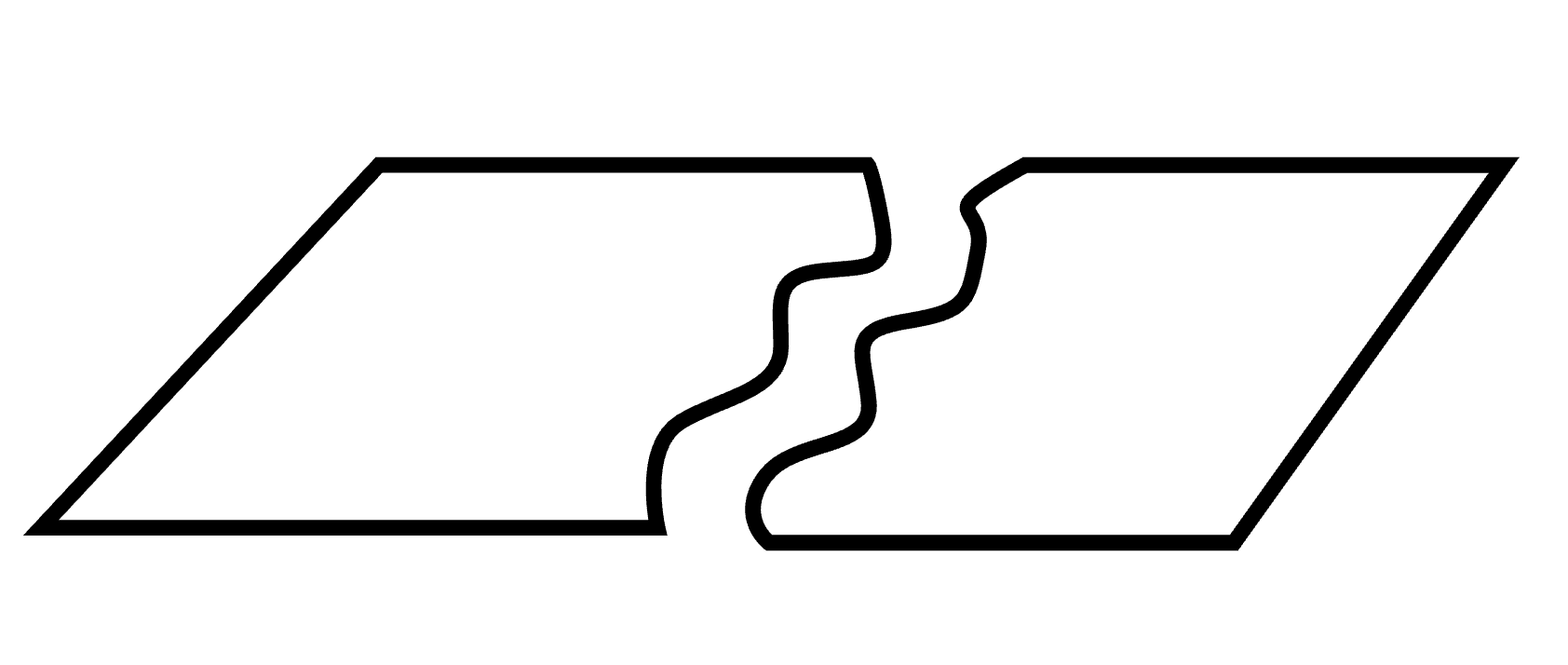}
        \caption{Simple disconnection}
    \end{subfigure}
    \begin{subfigure}[b]{0.24\textwidth}
        \centering
        \includegraphics[width=\linewidth]{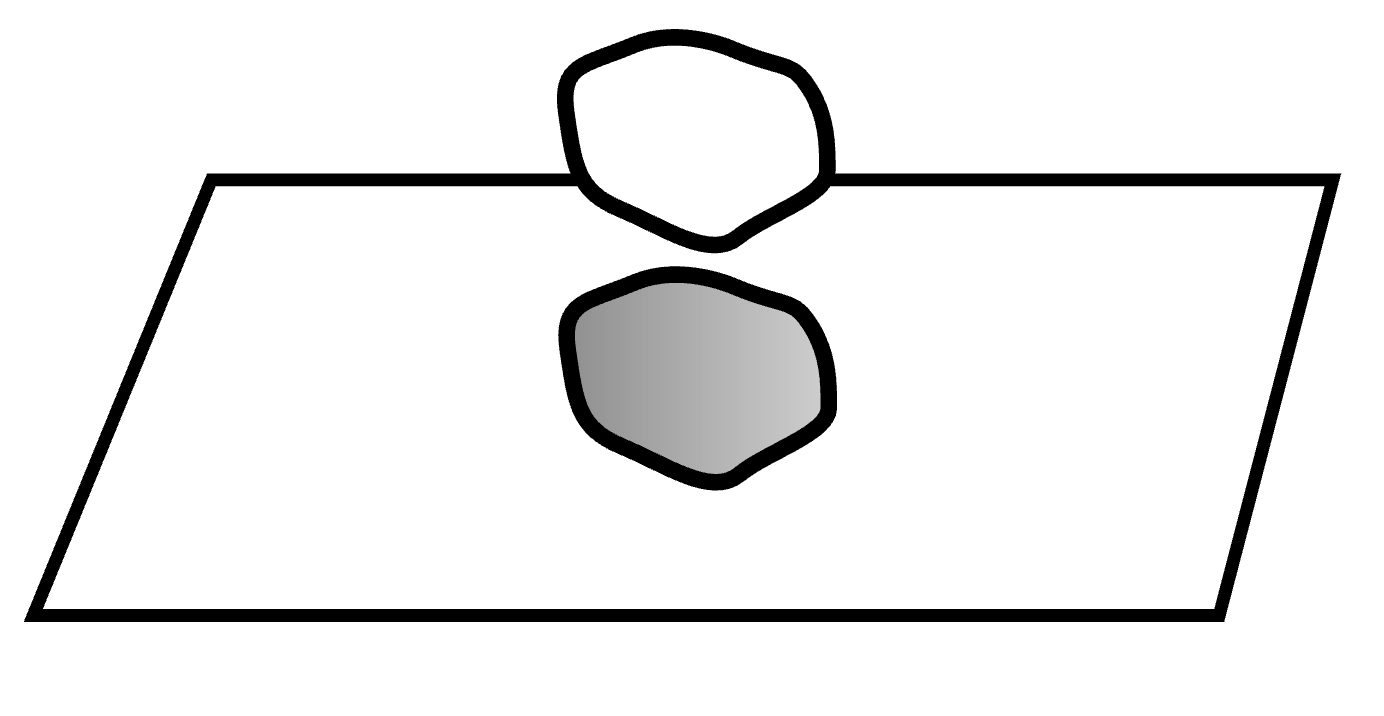}
        \caption{Hole disconnection}
    \end{subfigure}
    \caption{Schematic visualization of the four ETCs. In each case, the torn range $\Phi(\mathcal{M}\setminus\Gamma)$ is shown, representing isometric deformation of a flat sheet after tearing. The partial tears (a) and (b) leave the torn range connected, while cases (c) and (d) result in two connected components. In (d), the smaller, elevated component is homeomorphic to a disk, while the larger component is homeomorphic to an annulus.}
    \label{fig:elementary-topological-changes}
\end{figure*}

%% file: figures/t-SfT.tex
\begin{figure}
    \centering

\begin{tikzpicture}[
            scale=1.2,
            arrow/.style={-{Latex}, thick}
        ]

\draw[dashed] (1.5,-0.9) to (1.5,3);

\node at (-2,0) {
    \begin{tikzpicture}
        \draw[fill=gray!50] (-1,-1) rectangle (1,1);
        \draw[color=red, dashed, line width=1pt] (0,-1) -- (0.8, 1);
    \end{tikzpicture}
};


\node at (-2, 3) {
    \begin{tikzpicture}[scale=0.45]
    \begin{axis}[ticks=none,hide axis, colormap={blueblack}{color=(gray!100) color=(gray!30)}]
        \addplot3[line width=1pt, samples=31, samples y=0,domain=-1:1,variable=\t]
            ({\t},{1},{1+sin(90)});
        \addplot3[line width=2pt, samples=31, samples y=0,domain=-1:1,variable=\t]
            ({-1},{\t},{1+sin(90*\t)^2});
        \addplot3[surf,
            opacity=0.9,
            domain=-1:1,
            shader = flat,
            shader=interp,
            samples=31,
        ] {1+sin(90*y)^2};
        \addplot3[line width=1pt, samples=31, samples y=0,domain=-1:1,variable=\t]
            ({1},{\t},{1+sin(90*\t)^2});
        \addplot3[line width=1pt, samples=31, samples y=0,domain=-1:1,variable=\t]
            ({\t},{-1},{1+sin(90)});
        \addplot3[line width=1pt, samples=31, samples y=0, domain=0:1, color=red, variable=\t,  dashed]
            ({\t*0.8},{\t*2-1},{1+sin(90*(\t*2-1))^2});
    \end{axis}
    \end{tikzpicture}
};

\node at (1.7, 3) {
    \begin{tikzpicture}[scale=0.5]
    \begin{axis}[ticks=none,hide axis, colormap={blueblack}{color=(gray!100) color=(gray!30)}]
        \addplot3[line width=2pt, samples=31, samples y=0,domain=-1:1,variable=\t] (
            {cos(60)*( ((1+1)/2 *(9/10 - 1/2)+1/2) * (\t+1) -1) - sin(60) * 1},
            {sin(60)*( ((1+1)/2 *(9/10 - 1/2)+1/2) * (\t+1) -1) + cos(60) * 1},
            {3+1+0.8*sin(90 * 1)}
        );
        \addplot3[line width=2pt, samples=31, samples y=0,domain=-1:1,variable=\t] (
            {cos(60)*( ((-1+1)/2 *(9/10 - 1/2)+1/2) * (\t+1) -1) - sin(60) * -1},
            {sin(60)*( ((-1+1)/2 *(9/10 - 1/2)+1/2) * (\t+1) -1) + cos(60) * -1},
            {3+1+0.8*sin(90 * -1)}
        );
        \addplot3[surf,
            opacity=0.9,
            domain=-1:1,
            shader=interp,
            samples=31,
        ] (
        {cos(60)*( ((y+1)/2 *(9/10 - 1/2)+1/2) * (x+1) -1) - sin(60) * y},
        {sin(60)*( ((y+1)/2 *(9/10 - 1/2)+1/2) * (x+1) -1) + cos(60) * y},
        {3+1+0.8*sin(90 * y)}
        );
        \addplot3[line width=1pt, samples=31, samples y=0,domain=-1:1,variable=\t] (
            {cos(60)*( ((\t+1)/2 *(9/10 - 1/2)+1/2) * (-1+1) -1) - sin(60) * \t},
            {sin(60)*( ((\t+1)/2 *(9/10 - 1/2)+1/2) * (-1+1) -1) + cos(60) * \t},
            {3+1+0.8*sin(90 * \t)}
        );
        \addplot3[line width=1pt, samples=31, samples y=0,domain=-1:1,variable=\t, color=red, dashed] (
            {cos(60)*( ((\t+1)/2 *(9/10 - 1/2)+1/2) * (1+1) -1) - sin(60) * \t},
            {sin(60)*( ((\t+1)/2 *(9/10 - 1/2)+1/2) * (1+1) -1) + cos(60) * \t},
            {3+1+0.8*sin(90 * \t)}
        );
        \addplot3[line width=1pt, samples=31, samples y=0,domain=-1:1,variable=\t] (
            {((-1+1)/2 *(1/10 - 1/2)+1/2) * (\t+1) + 1},
            {-1},
            {3+0.5*(-1-\t)}
        );
        \addplot3[surf,
            opacity=0.9,
            domain=-1:1,
            shader=interp,
            samples=31,
        ] (
            {((y+1)/2 *(1/10 - 1/2)+1/2) * (x+1) + 1},
            {y},
            {3+0.5*(y-x)}
        );
        \addplot3[line width=1pt, samples=31, samples y=0,domain=-1:1,variable=\t] (
            {((1+1)/2 *(1/10 - 1/2)+1/2) * (\t+1) + 1},
            {1},
            {3+0.5*(1-\t)}
        );
        \addplot3[line width=1pt, samples=31, samples y=0,domain=-1:1,variable=\t] (
            {((\t+1)/2 *(1/10 - 1/2)+1/2) * (-1+1) + 1},
            {\t},
            {3+0.5*(\t-(-1)}
        );
        \addplot3[line width=1pt, samples=31, samples y=0,domain=-1:1,variable=\t, color=red, dashed] (
            {((\t+1)/2 *(1/10 - 1/2)+1/2) * (1+1) + 1},
            {\t},
            {3+0.5*(\t-1)}
        );

        \addplot3[only marks, mark=*, black] coordinates {(-0.39,0.25,3)};
    \end{axis}
    \end{tikzpicture}
};

\node at (2, 0) {
    \begin{tikzpicture}[scale=0.6]
    \begin{axis}[ticks=none,hide axis, colormap={blueblack}{color=(gray!100) color=(gray!30)}]
        \addplot3[only marks, mark=*, blue] coordinates {(0,0,0)};
        \draw[dashed,blue] (axis cs:0,0,0) -- (axis cs:-0.4,-0.8,1) ;
        \draw[dashed,blue] (axis cs:0,0,0) -- (axis cs:1.2,-0.8,1) ;
        \draw[dashed,blue] (axis cs:0,0,0) -- (axis cs:-0.4,0.6,1) ;
        \draw[dashed,blue] (axis cs:0,0,0) -- (axis cs:1.2,0.6,1) ;

        \addplot3[surf,
            opacity=0.4,
            domain=-0.4:1.2,
            y domain=-0.8:0.6,
            shader=interp,
            samples=31,
        ] (
            {x},
            {y},
            {1}
        );
        
        \addplot3[line width=2pt, samples=31, samples y=0,domain=-1:1,variable=\t] (
            {(cos(60)*( ((1+1)/2 *(9/10 - 1/2)+1/2) * (\t+1) -1) - sin(60) * 1 ) / (3+1+0.8*sin(90 * 1))},
            {(sin(60)*( ((1+1)/2 *(9/10 - 1/2)+1/2) * (\t+1) -1) + cos(60) * 1) / (3+1+0.8*sin(90 * 1))},
            {1}
        );
        \addplot3[line width=2pt, samples=31, samples y=0,domain=-1:1,variable=\t] (
            {(cos(60)*( ((-1+1)/2 *(9/10 - 1/2)+1/2) * (\t+1) -1) - sin(60) * -1) / (3+1+0.8*sin(90 * -1))},
            {(sin(60)*( ((-1+1)/2 *(9/10 - 1/2)+1/2) * (\t+1) -1) + cos(60) * -1) / (3+1+0.8*sin(90 * -1))},
            {1}
        );
        \addplot3[surf,
            opacity=0.9,
            domain=-1:1,
            shader=interp,
            samples=31,
        ] (
            {(cos(60)*( ((y+1)/2 *(9/10 - 1/2)+1/2) * (x+1) -1) - sin(60) * y) / (3+1+0.8*sin(90 * y))},
            {(sin(60)*( ((y+1)/2 *(9/10 - 1/2)+1/2) * (x+1) -1) + cos(60) * y) / (3+1+0.8*sin(90 * y))},
            {1}
        );
        \addplot3[line width=1pt, samples=31, samples y=0,domain=-1:1,variable=\t] (
            {(cos(60)*( ((\t+1)/2 *(9/10 - 1/2)+1/2) * (-1+1) -1) - sin(60) * \t) / (3+1+0.8*sin(90 * \t))},
            {(sin(60)*( ((\t+1)/2 *(9/10 - 1/2)+1/2) * (-1+1) -1) + cos(60) * \t) / (3+1+0.8*sin(90 * \t))},
            {1}
        );
        \addplot3[line width=1pt, samples=31, samples y=0,domain=-1:1,variable=\t, color=red, dashed] (
            {(cos(60)*( ((\t+1)/2 *(9/10 - 1/2)+1/2) * (1+1) -1) - sin(60) * \t) / (3+1+0.8*sin(90 * \t))},
            {(sin(60)*( ((\t+1)/2 *(9/10 - 1/2)+1/2) * (1+1) -1) + cos(60) * \t) / (3+1+0.8*sin(90 * \t))},
            {1}
        );
        \addplot3[line width=1pt, samples=31, samples y=0,domain=-1:1,variable=\t] (
            {(((-1+1)/2 *(1/10 - 1/2)+1/2) * (\t+1) + 1) / (3+0.5*(-1-\t))},
            {-1 / (3+0.5*(-1-\t))},
            {1}
        );
        \addplot3[surf,
            opacity=0.9,
            domain=-1:1,
            shader=interp,
            samples=31,
        ] (
            {(((y+1)/2 *(1/10 - 1/2)+1/2) * (x+1) + 1) / (3+0.5*(y-x))},
            {y / (3+0.5*(y-x))},
            {1}
        );
        \addplot3[line width=1pt, samples=31, samples y=0,domain=-1:1,variable=\t] (
            {(((1+1)/2 *(1/10 - 1/2)+1/2) * (\t+1) + 1) / (3+0.5*(1-\t))},
            {1 / (3+0.5*(1-\t))},
            {1}
        );
        \addplot3[line width=1pt, samples=31, samples y=0,domain=-1:1,variable=\t] (
            {(((\t+1)/2 *(1/10 - 1/2)+1/2) * (-1+1) + 1) / (3+0.5*(\t-(-1))},
            {\t / (3+0.5*(\t-(-1))},
            {1}
        );
        \addplot3[line width=1pt, samples=31, samples y=0,domain=-1:1,variable=\t, color=red, dashed] (
            {(((\t+1)/2 *(1/10 - 1/2)+1/2) * (1+1) + 1) / (3+0.5*(\t-1))},
            {\t / (3+0.5*(\t-1))},
            {1}
        );

        \addplot3[only marks, mark=*, black] coordinates {(0.075,-0.13,1)};
    \end{axis}
    \end{tikzpicture}
};

\draw[dashed] (1.5,0.33) to (1.5,1);

\node (P) at (-2.5,0.5) {$\mathcal{P}$};
\node (T) at (-2.3,3.5) {$\mathcal{T}$};
\node (S) at (1.2,3.5) {$\mathcal{S}$};
\node (J) at (3,0.4) {$\mathcal{J}$};

\draw[arrow] (-2.2, 1) to[bend left=10] node[left] {$\Delta$} (-2.2,2.2);
\draw[arrow] (-0.7, 3) to[bend left=10] node[above] {$\Phi$} (1,3);
\draw[arrow] (-1, -0.1) to[bend right=10] node[above] {$\eta$} (0.5,-0.1);
\draw[arrow] (2, 2.2)  to node[right] {$\Pi$} (2, 0.9);
\draw[arrow] (-1.3, 1) to[bend left = 10] node[above] {$\varphi$} (1,2.9);
\draw[arrow] (-1, 0.7) to[bend left = 10] node[above] {$\gamma$} (1.4,1.5);

\node (GammaLine) at (-1.64, -0.14) {};
\node (GammaLabel) at (-2.1, -0.1) {$\Gamma$};
\draw[->] (GammaLabel) -- (GammaLine);

\end{tikzpicture}

    \caption{\y{sft} \textit{with} topological changes. The formalism is as in \cref{fig:SfT}, but now 
    the unknown is an isometry with topological change $\Phi$ with tearing curve $\Gamma.$ In particular, $\Phi$ is a simple disconnection \y{etc}.}
    \label{fig:t-SfT}
\end{figure}
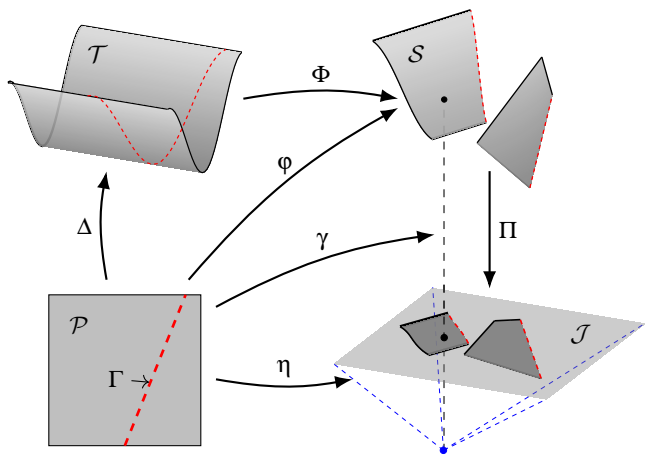

%% file: sections/results.tex
In this section, we present the experimental evaluation of our method. We first describe the datasets followed by quantitative comparisons with representative baseline approaches. We also provide qualitative results to illustrate the behavior of our method under diverse conditions, including cases involving significant topological changes. Finally, we analyze the results, highlighting both its robustness across scenarios and the aspects that remain challenging.

\subsection{Compared methods}
We benchmarked our approach against seven established \y{sft} methods for the \y{sft} problem. Specifically, we included \textit{BGCC12}, the closed-form local isometric solution of \cite{bartoli2012template}; \textit{BGCC12-R}, which extends \textit{BGCC12} with the non-linear refinement procedure of \cite{brunet2014monocular}; \textit{CPB14}, the normal-integration approach of \cite{chhatkuli2014stable}; \textit{CPB14-R}, the refinement of \textit{CPB14} following \cite{brunet2014monocular}; \textit{BHB11}, the second-order cone programming formulation of \cite{brunet2011monocular}; \textit{PPB19}, the Cartan’s formulation–based method of \cite{parashar2019local} which is further split into \textit{PPB19-IsoConf} and \textit{PPB19-Smooth} for its isometric-conformal and smooth deformation-based variants respectively; and \textit{SAOB24}, a non-real-time adaptation of the method proposed in \cite{shetab2024robusft}.

\subsection{Evaluation metrics}
The metric used for evaluating accuracy is \y{rmse}, which is given by: 
\begin{equation}
    \text{RMSE} = \sqrt{\frac{1}{n}\sum_{i=1}^n \| \mathbf{p}_{i, \text{reconstruction}} - \mathbf{p}_{i, \text{ground truth}}\| ^2 }.
\end{equation}
\noindent Since the input data consistently provides keypoint correspondences, we are able to employ \y{rmse} as a universal accuracy measure. This ensures a direct, point-to-point comparison between estimated and \y{gt} configurations, and eliminates the need for correspondence-free metrics such as the Hausdorff or Chamfer distances.

\subsection{Dataset}
\input{tables/algo}
\begin{figure}[t]
    \centering
    \begin{overpic}[width=\columnwidth]{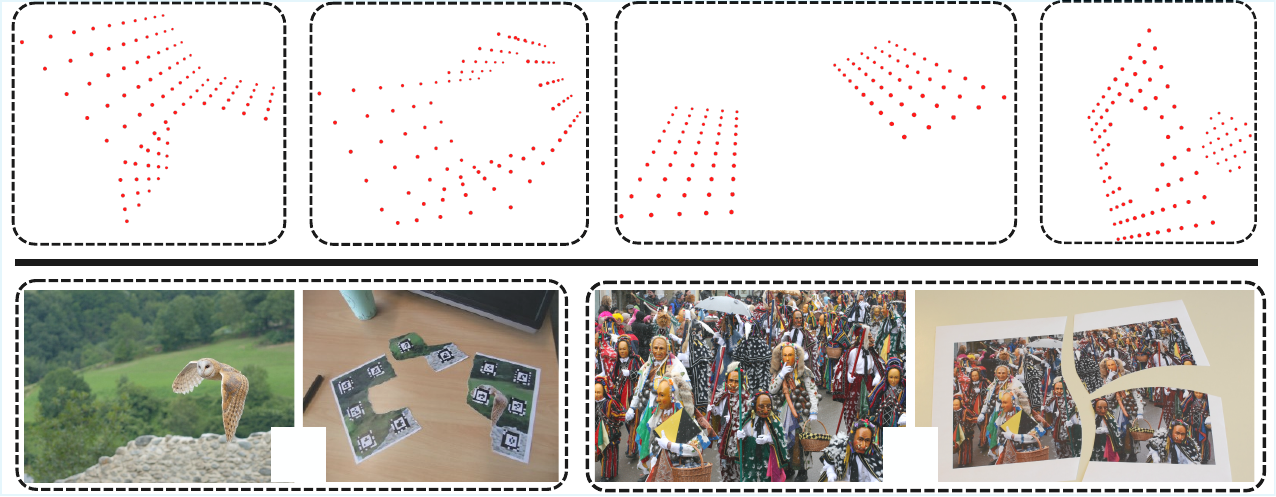}
        \put(15,22){(a)}
        \put(26,35){(b)}
        \put(62,22){(c)}
        \put(83,35){(d)}
        \put(21.5,2){\footnotesize{(e)}}
        \put(69.5,2){\footnotesize{(f)}}
    \end{overpic}
    \caption{Dataset used in the experiments. (a)--(d): synthetic \yp{etc};  (e) and (f): torn papers.}
    \label{fig:images}
\end{figure}
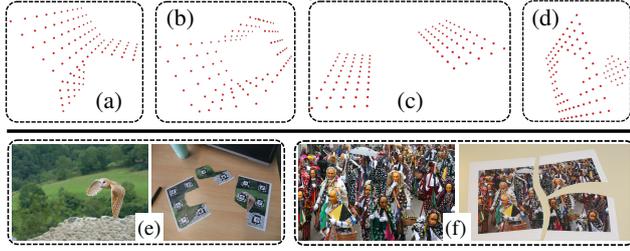
We validate our method on both synthetic and real data, as described below. All datasets given below use as a template a rectangular un-torn plane of data-appropriate dimensions.

\noindent \textbf{Elementary topological changes}.~We generate synthetic data, corresponding to the four elementary topological changes described in \cref{fig:elementary-topological-changes}, by simulating 100 keypoints in 3D and using arbitrary known intrinsics and unknown extrinsics to project the data to complete the \y{sft} setup. The four elementary topological changes were simulated according to the following parametric descriptions of surfaces, where the domain is $(s,t)\in [-1,1]^2\setminus\Gamma$. 
\begin{enumerate}[(a)]
    \item Exterior tear: $\Gamma = \{0\}\times [0,1]$.
    \[ 
    \begin{cases}
        (s,t,2) & t\leq 0, s\neq 0 \\
        (s, (\sin 0.8) t, 2+(\cos 0.8)t) & t>0, s<0 \\
        (s, (\sin 0.8) t, 2-(\cos 0.8)t) & t>0, s> 0
    \end{cases}
    \]
     \item Interior tear: $\Gamma = \{0\}\times [-1,1]$. 
    \[
    \begin{cases}
        (s, \cos(0.75)t, 3+\sin(0.75)t)  & t\leq 0, s>0 \\
        (s, \cos(0.75)t, 3-\sin(0.75)t) &  t>0, s>0 \\
        (s, \cos(0.75)t, 3-\sin(0.75)(2+t)) &  t\leq 0, s<0 \\
        (s, \cos(0.75)t, 3-\sin(0.75)(2-t)) &  t>0, s<0\\
    \end{cases}
    \]
    \item Simple disconnection: $\Gamma = [-1,1]\times\{0\}$, and $Q_1$, $Q_2$ are transformations in $\mathbb{SE}(3)$
    \[
    \begin{cases}
        Q_1(s-0.5, t, 1) & s<0 \\
        Q_2(s+0.5,t,1) & s>0\\
    \end{cases}
    \]
    \item Hole disconnection: $\Gamma = \{\sqrt{s^2+t^2} = 3/5\}$ and $Q$ is a transformation in $\mathbb{SE}(3)$.
    \[
    \begin{cases}
        Q(s,t,1) & \sqrt{s^2+t^2} > 3/5 \\
        Q(s,t,3) & \sqrt{s^2+t^2} < 3/5 \\
    \end{cases}
    \]
\end{enumerate}

\noindent All values in this dataset are in \y{au}.


\noindent \textbf{Torn papers}.~We use two real data of papers torn apart arbitrarily and captured with a calibrated camera. Two distinct textures were used, the first texture is of a barn owl in flight over the Pyrénées\footnote{\tiny{commons.wikimedia.org/wiki/File:Tyto\_alba\_1\_Luc\_Viatour.jpg}}, we call it the \textit{owl data}, and the second texture is from the Swabian-Alemannic Fastnacht\footnote{\tiny{de.wikipedia.org/wiki/Datei:Fastnachtsumzug,\_Rottweil,\_S\%C3\%BCddeutschland.jpg}} festival, called the \textit{SAF data}. The textures are printed on paper, torn apart and photoed while correspondences between the template and the captured image are established using RoMa~\cite{edstedt2024roma}. The \y{gt} was obtained by manually annotating the separation boundaries, which yields a clean tearing-aware clustering of matched features; \y{sft} was then run independently on each torn piece of paper. Since modern \y{sft} approaches (\textit{PPB19} was used specifically) achieve high accuracy in the absence of topological change, they are assumed to provide the \y{gt} within each torn piece. All values in this dataset are in meters.

Both datasets have been visualized in \cref{fig:images}, the \y{gt} keypoints for \yp{etc} and the input template and image for torn papers.

\subsection{Description of Results}
We describe the results from our experiments below.

\noindent \textbf{\y{etc}}.~We tabulate the \y{rmse} for all benchmark methods as well as our proposed method for the \y{etc} dataset in \Cref{tab:synth_res}. We visualize the reconstruction for all methods on \y{etc} in \cref{fig:qual_etc}.

On the \y{etc} dataset, our method achieves the best performance across all cases, with the sole exception of the \textit{interior tear} scenario, where \textit{BGCC12} attains an \y{rmse} that is 8.33\% lower than ours. When results over all four \y{etc} instances are aggregated, however, \textit{BGCC12} is neither the second nor the third-best performing method; those positions are instead occupied by \textit{PPB19-IsoConf} and \textit{PPB19-Smooth}, respectively. Overall, our method yields a mean accuracy improvement of 15\% relative to the next-best approach, \textit{PPB19-IsoConf}. Moreover, it is important to note that neither \textit{PPB19-IsoConf} nor \textit{PPB19-Smooth} is able to successfully reconstruct all four \y{etc} instances, underscoring the broader applicability of our framework. 

\input{tables/syntheticResultsTable}




\begin{figure}[t]
    \centering
    \begin{subfigure}[t]{0.48\columnwidth}
        \centering
        \includegraphics[width=\linewidth]{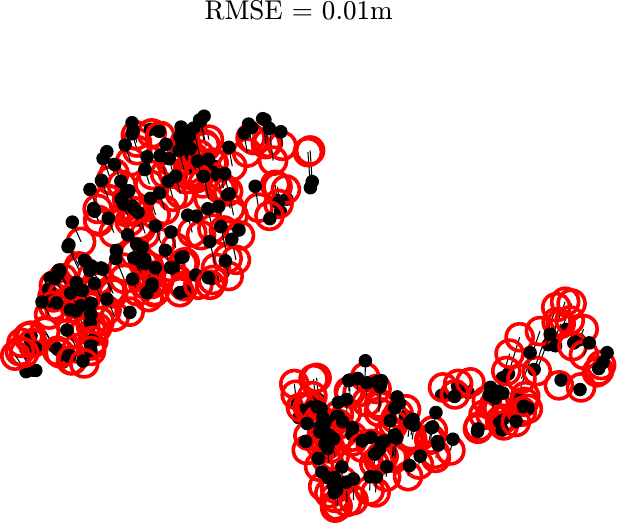}
        \subcaption{}
    \end{subfigure}\hfill
    \begin{subfigure}[t]{0.48\columnwidth}
        \centering
        \includegraphics[width=\linewidth]{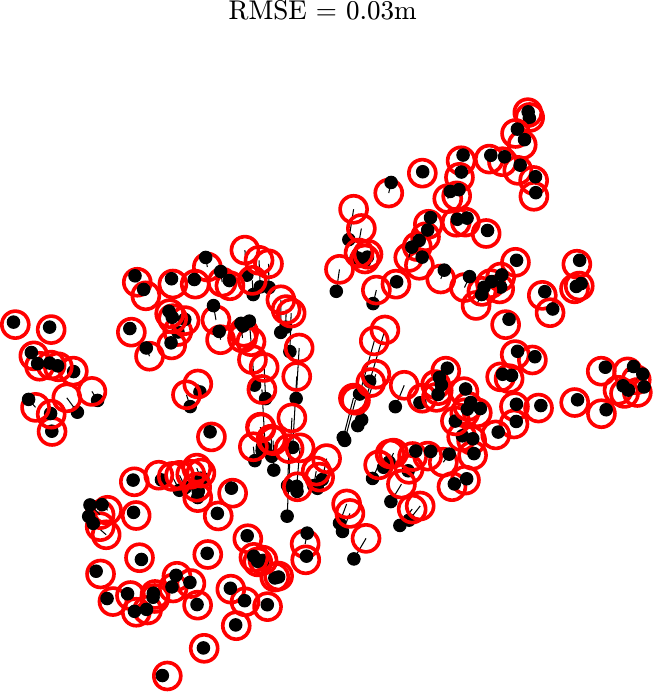}
        \subcaption{}
    \end{subfigure}
    \caption{Reconstruction results from (a)~\textit{owl data} and (b)~\textit{SAF data}; the red hollow points are \y{gt} and the black dots are the reconstructions from our proposed method. The corresponding reconstructed and \y{gt} points are connected by think black lines.}
    \label{fig_real_qual}
\end{figure}

\noindent\textbf{Torn paper}.~In the \textit{owl data}, we pick 270 high-confidence feature matches from RoMa~\cite{edstedt2024roma} for validating our proposed approach. These features are clustered around the two large torn pieces in the captured image, the third small torn piece did not have any correspondences, thus being eliminated from \y{sft}. The reconstruction results are highly accurate, given the large separation boundary -- the achieved \y{rmse} is 1.3~$cm$. The results have been qualitatively visualized in \cref{fig_real_qual}a. In the \textit{SAF data}, we pick 197 high-confidence feature matches from RoMa~\cite{edstedt2024roma} for validating our proposed approach. The features are well-distributed among the three torn pieces. We achieve a reconstruction accuracy of 3.3~$cm$ in terms of \y{rmse}. The results have been qualitatively visualized in \cref{fig_real_qual}b.

\subsection{Discussion}
We highlight below the strengths and challenges of our proposed method along with details of the runtime.

\noindent\textbf{Strengths}.~The strengths of our approach are most evident in the case of hole disconnections. Such tears are particularly challenging because they often induce substantial geometric distortions without large-scale separations of connected components in the perspective projection. It is therefore consistent with expectation that existing methods, which do not explicitly account for topological changes, struggle to achieve accurate reconstructions in these settings, whereas our method is more accurate, yielding a 46.7\% gain over \textit{BGCC12}, while \textit{PPB18-IsoConf} and \textit{PPB18-Smooth} fail to converge.


\noindent\textbf{Challenges}.~Interestingly, \textit{BGCC12}, \textit{PPB18-IsoConf}, and \textit{PPB18-Smooth} achieve reconstructions comparable to ours and in the case of interior tearing, \textit{BGCC12} even slightly outperforms, despite none of these methods being designed to handle topological changes explicitly. For exterior and interior tears, this can be attributed to the limited extent of tearing and deformation (\cref{fig:qual_etc}) which remains within the noise tolerance typically accommodated by \y{sft} formulations. Thus, the results for exterior and interior tears highlight the inherent robustness of \textit{BGCC12}, \textit{PPB18-IsoConf}, and \textit{PPB18-Smooth}, and not their adaptability to topological changes. The incompatibility between data and model assumptions become evident in \textit{BGCC12-R}, where a non-linear refinement of \textit{BGCC12} causes a sharp decline in accuracy; an outcome unlikely under model-data consistency. For simple disconnection, all methods exhibit noticeable degradation, with the best \y{rmse} reaching only 0.43 compared to 0.15 and 0.11 for exterior and interior tears, respectively. As our method is initialized from classical \y{sft} solutions, it shares similar limitations, yet consistently refines the estimates to marginally surpass competing approaches. 

\noindent \textbf{Runtime}.~Our current Python implementation has not yet been optimized for efficiency and therefore operates only in a batch-processing mode; for example, the \y{etc} dataset requires on average 913.5 seconds on an Apple M1 Pro with 16GB of RAM. While our current Python prototype is not optimized, we note that the core operations (radial basis interpolation, gradient descent updates) are highly parallelizable on GPU, and we expect an order-of-magnitude speedup with a compiled/CUDA implementation.

%% file: tables/algo.tex
\begin{algorithm}[b]
\caption{Topological-change-aware \y{sft}}
\label{alg:topo-sft}
\begin{algorithmic}[1]
\Require Templates $(\mathcal{P},\mathcal{T})$, image domain $\mathcal{J}$, normalized keypoint correspondences 
$\{(\mathbf{p}_i,\mathbf{p}'_i)\}_{i=1}^m$ with $\mathbf{p}_i\in\mathcal{P}$ and $\mathbf{p}'_i\in\mathcal{J}$
\Ensure Reconstructed 3D points $\{\mathbf{P}_i\}_{i=1}^m$

\State \textbf{Warp estimation:} 
Compute template-to-image warp $\eta:\mathcal{P}\to\mathcal{J}$ 
and template parametrization $\Delta:\mathcal{P}\to\mathcal{T}$ 
from correspondences.

\State \textbf{Initial reconstruction:} 
Compute depth function $\gamma_0:\mathcal{P}\to\mathbb{R}$ via Eq.~(3) and form
\[
\Phi_0 = \varphi_0\circ \Delta^{-1}, \qquad 
\varphi_0(\mathbf{p}) = \widetilde\eta(\mathbf{p})\,\gamma_0(\mathbf{p}).
\]

\State \textbf{Refinement by displacement field:} 
Introduce displacement field $d:\mathcal{P}\to\mathbb{R}^2$ 
and define the refined reconstruction
\[
\Phi_d(\Delta(\mathbf{p})) = \widetilde\eta(\mathbf{p})\,\gamma_0(\mathbf{p}+d(\mathbf{p})).
\]

\State \textbf{Optimization:} 
Determine $d$ by minimizing the discrete isometric error function in \cref{eq:cost}. Do gradient descent until convergence.

\State \textbf{Output:} 
Return reconstructed 3D correspondences \[\mathbf{P}_i=\Phi_d(\mathbf{p}_i)\]
\end{algorithmic}
\end{algorithm}

%% file: tables/syntheticResultsTable.tex
\begin{table*}[t] \begin{adjustbox}{width=\textwidth,center} \begin{tabular}{rccccccccc}  \toprule \hline & \cellcolor[RGB]{200,200,200} \textbf{Ours} & \cellcolor[RGB]{200,200,200} \textbf{BGCC12} & \cellcolor[RGB]{200,200,200} \textbf{BGCC12-R} & \cellcolor[RGB]{200,200,200} \textbf{CPB14} & \cellcolor[RGB]{200,200,200} \textbf{CPB14-R} & \cellcolor[RGB]{200,200,200} \textbf{BHB11} & \cellcolor[RGB]{200,200,200} \textbf{PPB19-IsoConf} & \cellcolor[RGB]{200,200,200} \textbf{PPB19-Smooth} & \cellcolor[RGB]{200,200,200} \textbf{SAOB24}\\ \cline{2-10} \textbf{Exterior tear}  & \textbf{\underline{0.15}} &\textbf{0.17} &0.36 &0.46 &0.48 &0.73 &0.19 &\ding{56} &0.34\\\textbf{Interior tear}  & \textbf{0.12} &\textbf{\underline{0.11}} &0.48 &0.34 &0.39 &0.36 &0.21 &0.22 &0.3\\\textbf{Simple disconnection}  & \textbf{\underline{0.43}} &0.48 &1.22 &\textbf{0.46} &0.85 &1.52 &0.8 &0.72 &1.66\\\textbf{Hole disconnection}  & \textbf{\underline{0.66}} &1.24 &1.4 &1.22 &1.27 &\textbf{0.86} &\ding{56} &\ding{56} &1.71\\\hline \textbf{Combined}  & \textbf{\underline{0.34}} &0.5 &0.87 &0.62 &0.75 &0.87 &\textbf{0.4} &0.47 &1\\\hline\bottomrule \end{tabular} \end{adjustbox} \caption{\y{rmse} of reconstruction by our proposed approach (first column) and the baseline methods. The best method for each data is marked with \underline{\textbf{bold-underscore}}, the second-best is \textbf{bold}, and the methods that failed to produce a valid result are marked with \ding{56}. All values are given in \y{au}.}   \label{tab:synth_res} \end{table*}

%% file: sections/conclusion.tex
\begin{figure*}[btp]
    \centering
    \begin{overpic}[width=0.7\textwidth]{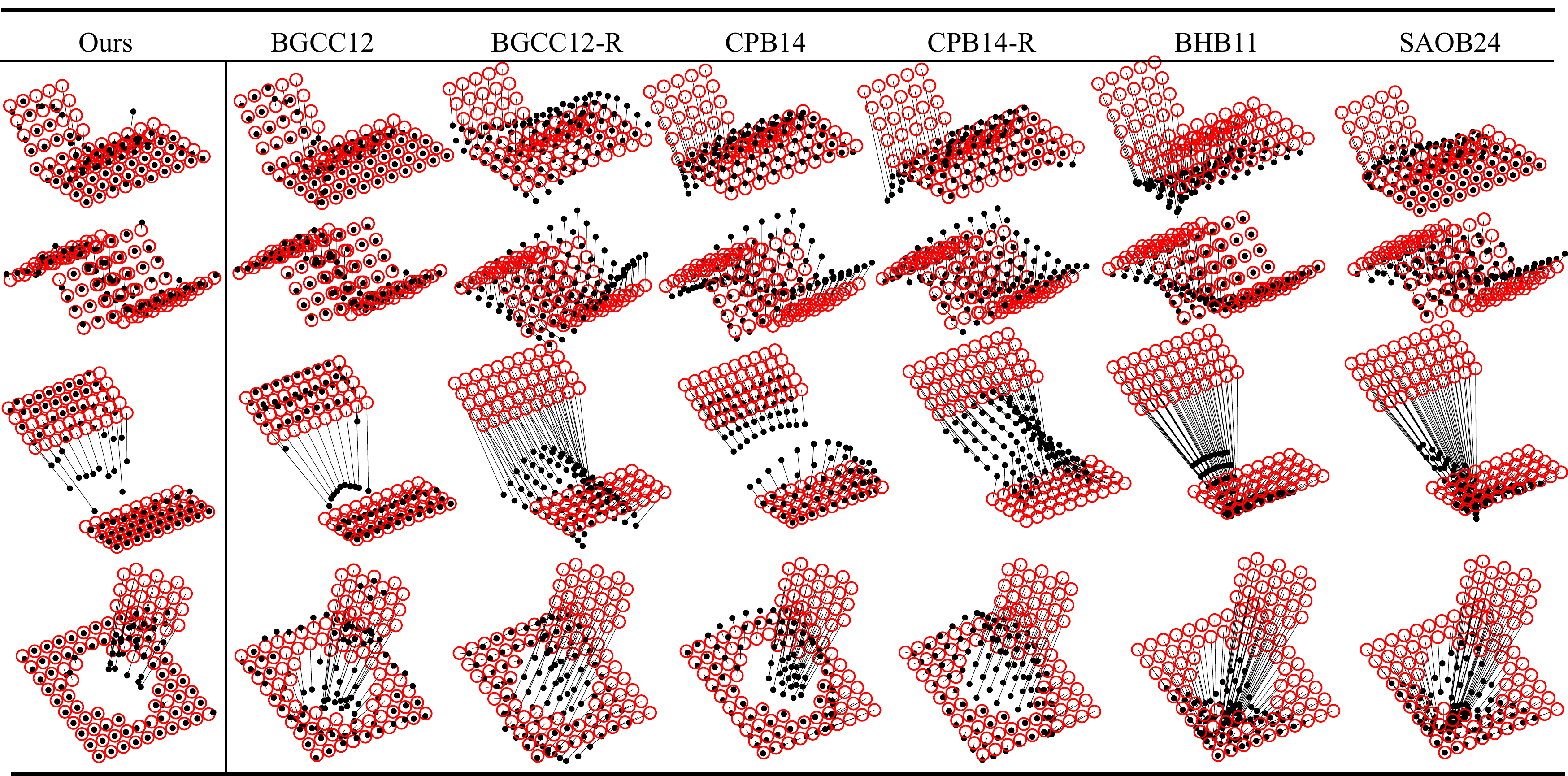}
    \end{overpic}
    \caption{Results obtained from synthetic \y{etc}; red hollow points are \y{gt}, black dots are the reconstructions from the algorithms mentioned on the table header and the corresponding \y{gt} and reconstructed points are connected via thin black lines -- we visualize these qualitative results for all methods that successfully produced results on all four \yp{etc}.}
    \label{fig:qual_etc}
\end{figure*}

As \textit{future work}, we propose two directions of investigation for improving the applicability of our method. The first idea is to modify the kernel of the warp $\eta$ as part of the optimization problem such that the displacement information updates distances between points in the kernel. This requires development of a tearing adapted radial basis function -- a non-trivial endeavor. The second idea is to seed the reconstruction from a small patch of the observed surface and investigate strategies for growing this patch in a manner such that all the torn pieces are reconstructed sequentially. 

In \textit{conclusion}, we have proposed a method to reconstruct objects undergoing topological changes from 2D images captured perspectively. The problem was challenging with no previously known solutions. We have formalized the notion of topological changes in the context of surface representations used in \y{sft}. The proposed method has been validated on multiple synthetic and real data and compared with state-of-the-art approaches showing our improved accuracy and applicability. Incorporating our method into real-world pipelines (e.g., surgical \y{ar}, robotic cutting, reconstruction of garments) is a natural next step.

